\documentclass[2025,preprint,12pt]{elsarticle}

\usepackage[pdfborder={0 0 0}]{hyperref}
\usepackage{amsthm,amsmath,amsfonts,latexsym,amssymb}
\usepackage{graphics,fancyhdr,graphicx}
\usepackage[margin=1in]{geometry}
\usepackage{multirow,booktabs}
\usepackage{xcolor}
\usepackage{listings}
\usepackage{tabularx}
\usepackage{placeins}
\usepackage{comment}
\usepackage{lineno}
\usepackage{lscape}
\usepackage{bm}
\usepackage{hyperref}
\usepackage{mathrsfs}
\usepackage{arydshln}
\usepackage{appendix}

\definecolor{custom-blue}{RGB}{3,69,173}
\hypersetup{
    colorlinks = true,
    urlcolor   = blue,
    citecolor  = custom-blue,
}

\biboptions{numbers,sort&compress}
\definecolor{listinggray}{gray}{0.9}
\definecolor{lbcolor}{rgb}{0.9,0.9,0.9}
\definecolor{Darkgreen}{RGB}{0,100,0}
\usepackage{caption}

\usepackage{float}
\usepackage{booktabs}
\usepackage{multirow}
\usepackage{tabularray}
\usepackage{mathtools}
\usepackage{subcaption}
\usepackage{enumitem}
\usepackage{rotating}
\usepackage{booktabs}
\usepackage{array}
\usepackage{hyperref}
\usepackage{graphicx}
\usepackage{comment}

\usepackage[margin=20pt]{subcaption}
\usepackage{float}
\usepackage{soul}
\usepackage{algorithm}
\usepackage{algpseudocode}
\usepackage{array,booktabs,arydshln,xcolor,ragged2e}

\usepackage{xcolor,colortbl}
\usepackage{multicol}
\usepackage{wrapfig}

\DeclarePairedDelimiter{\norm}{\lVert}{\rVert}
\DeclareMathOperator*{\argmax}{arg\,max}

\newcolumntype{C}[1]{>{\centering\arraybackslash}p{#1}}  

\begin{document}

\makeatletter
\def\ps@pprintTitle{%
  \let\@oddhead\@empty
  \let\@evenhead\@empty
  \let\@oddfoot\@empty
  \let\@evenfoot\@oddfoot
}
\makeatother

\abovedisplayskip=6.0pt
\belowdisplayskip=6.0pt

\begin{frontmatter}

\title{On Some Tunable Multi-fidelity Bayesian Optimization Frameworks}

\author[1]{Arjun Manoj
}
\author[1]{Anastasia S. Georgiou
}
\author[2]{Dimitris G. Giovanis
}
\author[3]{Themistoklis P. Sapsis
}
\author[1,4]{Ioannis G. Kevrekidis\corref{cor1}}
\ead{yannisk@jhu.edu}

\address[1]{Department of Chemical and Biomolecular Engineering, Johns Hopkins University, U.S.A.}
\address[2]{Department of Civil and Systems Engineering, Johns Hopkins University, U.S.A.}
\address[3]{Department of Mechanical Engineering, Massachusetts Institute of Technology, U.S.A.}
\address[4]{Department of Applied Mathematics and Statistics, Johns Hopkins University, U.S.A}

\cortext[cor1]{Corresponding author.}

\begin{abstract}
Multi-fidelity optimization employs surrogate models that integrate information from varying levels of fidelity to guide efficient exploration of complex design spaces while minimizing the reliance on (expensive) high-fidelity objective function evaluations. To advance Gaussian Process (GP)-based multi-fidelity optimization, we implement a proximity-based acquisition strategy that simplifies fidelity selection by eliminating the need for separate acquisition functions at each fidelity level. We also enable multi-fidelity Upper Confidence Bound (UCB) strategies by combining them with multi-fidelity GPs rather than the standard GPs typically used. We benchmark these approaches alongside other multi-fidelity acquisition strategies —including fidelity-weighted approaches— comparing their performance, reliance on high-fidelity evaluations, and hyperparameter tunability in representative optimization tasks. The results highlight the capability of the proximity-based multi-fidelity acquisition function to deliver consistent control over high-fidelity usage while maintaining convergence efficiency. Our illustrative examples include multi-fidelity chemical kinetic models, both homogeneous and heterogeneous (dynamic catalysis for ammonia production).
\end{abstract}


\end{frontmatter}


\section{Introduction}
\label{intro}

High-fidelity models are often required to optimize operations for dynamical systems characterized by complex dynamics in physicochemical or engineering contexts. However, such models \cite{sapsis, sapsis2, mfreactor, ferrari2024digital, kharazmi2021data} can be computationally expensive to evaluate, and their evaluation often constitutes the bottleneck in their use for optimization. 
The complexity levels and the lack of information on the {\em internal workings} of such models hinder the use of gradient-based optimization methods \cite{ei, BO, kelley2004newton} or brute-force search methods, and limit us to treating them as black-box functions. Active learning techniques like Bayesian Optimization (BO) \cite{BO, brochu2010tutorial} intelligently sample design spaces for such high-fidelity black-box models in the hopes of minimizing the number of expensive model evaluations made during optimization. 
A key component for BO is the {\em acquisition function}, which is constructed based on the surrogate model and guides the selection of the next best query location(s) during the active learning process. Traditional acquisition functions include improvement-based \cite{ei, pi, mockus1998application} and bandit-based strategies \cite{UCb, kaufmann2012bayesian}. More recent developments include information-based algorithms \cite{knowledgeGradient, villemonteix2009informational, hennig2012entropy, hernandez2014predictive, blanchard2021bayesian} which often exhibit performance advantages, especially for noisy functions, However, these strategies require substantial sampling and implementation effort, making it challenging to apply, especially in higher dimensions.

To further accelerate/improve the optimization scheme, we can also leverage  lower-fidelity model(s)  to infer useful information about the high-fidelity space while reducing the overall computational expense. Although low-fidelity models could be inherently noisy and approximate, they often capture key trends and qualitative behaviors which can be used to learn about the high-fidelity space. Combining several low-fidelity samples with fewer high-fidelity samples using multi-fidelity models \cite{mfgpr1, mfgpr2, popov2022multifidelity, perdikaris2017nonlinear} can facilitate optimization, often at a significantly reduced computational expense. Such multi-fidelity methods have been developed and applied in various contexts, including data assimilation \cite{popov2022multifidelity} for hierarchies of models and observations, nonlinear information fusion \cite{perdikaris2017nonlinear} for computational fluid dynamics simulations, composite neural networks that learn from multi-fidelity data \cite{meng2020composite, meng2021multi}, and reduced-order modeling frameworks using multi-fidelity long short-term memory (LSTM) networks \cite{conti2024multi} to predict the results partial differential equation (PDE) simulations. 

Multi-fidelity Gaussian Processes (GPs) \cite{mfgpr1, mfgpr2} extend standard GP regression by incorporating information from multiple sources of varying fidelity through hierarchical structures, allowing more accurate predictions and improved uncertainty quantification. Recent studies \cite{mfreactor, sapsis} highlight the use of multi-fidelity GPs as surrogate models for multi-fidelity systems. Sapsis et al. combined a multi-fidelity GP with a fidelity-weighted acquisition function \cite{sapsis}. The fidelity-weighted approach adjusts the acquisition function  using a cost-ratio penalty term to account for the significant difference in computational expense between the high-fidelity and low-fidelity functions. By appropriately adjusting this cost-ratio penalty, the acquisition function is biased toward selecting more points from the low-fidelity function. However, our preliminary investigations highlighted poor information exchange between low-fidelity and high-fidelity models in certain cases. Kandasamy et al. proposed an intuitive multi-fidelity acquisition function \cite{mfucb} based on Upper Confidence Bounds (UCB) which we sometimes find to exhibit improved information exchange. However, instead of a multi-fidelity GP, they utilize a standard GP as the high-fidelity surrogate model, constructed solely from the high-fidelity data. The low-fidelity model is also a standard GP trained using the low-fidelity data. Khatamsaz et. al \cite{arrojavemfbo} combined {\em reification} \cite{winkler1981combining} (as the multi-fidelity model) with the knowledge gradient \cite{knowledgeGradient} as the acquisition function. The so-called {\em Knowledge Gradient} they use involves the training of multiple GPs for a single iteration - a step that might be amenable to improvement. 

The overarching goal in this work is to develop and benchmark tunable Multi-Fidelity Bayesian Optimization (MFBO) frameworks aimed at minimizing reliance on high-fidelity evaluations while maintaining convergence efficiency and predictable hyperparameter control. In addition to integrating the multi-fidelity UCB acquisition function with multi-fidelity GPs, we develop a separate proximity-based acquisition function (with fewer parameters) to simplify fidelity selection. The fidelity selection is guided by the low-fidelity sample density in a proximity region defined using a hyperparameter based on the relative sampling costs of the different fidelity levels. Unlike the aforementioned strategies, our approach uses {\em a single acquisition function}, eliminating the need for separate acquisition functions for different fidelity levels.

We benchmark our framework across a variety of optimization problems, including (a) a simple enzyme reaction scheme~\cite{nikoenzyme} as well as a nonlinear chemical reaction model \cite{BZ-FKN2} in which the application of the Quasi Steady-State Approximation yields a low-fidelity model that is lower dimension, but only accurate in certain parameter regimes; and (b) a dynamic catalysis (periodically forced) $\text{NH}_3$ heterogeneous kinetic model~\cite{ammonia}, where varying the numerical integration tolerance creates two distinct fidelity models. We hope that these optimization problems, especially the state-of-the-art dynamic catalysis for commercial ammonia production (a key process in advancing renewable energy), highlight the potential of employing multi-fidelity frameworks to tackle the optimization of computationally intensive real-world problems.

\section{Methods}
\label{sec:methods}

\subsection{Bayesian Optimization}
\label{subsec:bo}
Bayesian Optimization (BO) optimizes computationally expensive functions by employing probabilistic surrogate models which -in addition-  estimate the associated uncertainty\cite{BO}. 
The algorithm actively learns by optimizing a computationally inexpensive acquisition function to obtain the next best design/evaluation conditions. This is a derivative-free approach that can be classified under black box optimization. Gaussian processes are a common choice for the probabilistic surrogate models for BO. We briefly revisit both Gaussian processes and acquisition functions in \ref{app:gp}.

\subsection{Multi-fidelity Gaussian Process Regression}

When several models -at different levels of fidelity- are available,  one can fruitfully combine them.   Using Multi-fidelity Gaussian Process Regression (MFGPR) we obtain hierarchical models by combining a linearly scaled low-fidelity Gaussian Process with a correction Gaussian Process \cite{mfgpr1, mfgpr2}. 
The correction GP provides flexibility in the modeling, allowing the high-fidelity model to improve the approximation of more complex behaviors. Although these multi-fidelity models can be scaled up to capture multiple (more than two) levels of fidelity, we only consider two levels in this work, as this aligns with the requirements of our test problems. The multi-fidelity GP for two levels of fidelity can be defined as

\begin{equation}
    Z_{low}(\bm{x}) \sim \mathcal{GP}(m_{low}(\bm{x}), \kappa_{low}(\bm{x}, \bm{x'}))
\end{equation}
\begin{equation}
    \delta(\bm{x}) \sim \mathcal{GP}(m_{\delta}(\bm{x}), \kappa_{\delta}(\bm{x}, \bm{x'}))
\end{equation}
\begin{equation}
    Z_{high}(\bm{x}) = \rho \cdot Z_{low}(\bm{x}) + \delta(\bm{x}),
\end{equation}
where,
\begin{itemize}
    \item $\rho$, a hyperparameter, is a scaling factor which weighs the low-fidelity contribution to the high-fidelity,
    \item $Z_{low}(\bm{x})$ is the low-fidelity GP,
    \item $\delta(\bm{x})$ is the correction GP,
    \item $Z_{high}(\bm{x})$ is the high-fidelity GP.
\end{itemize}
The low-fidelity GP ($Z_{low}$) is built using the low-fidelity data points, while the correction GP ($\delta$) is constructed based on the discrepancy (error) between the low-fidelity and high-fidelity data points. Both components are initialized as zero-mean GPs: for example, $Z_{low}(\bm{x}) \sim \mathcal{GP}(m_{low}=0, \kappa_{low}(\bm{x}, \bm{x'}))$, where $\kappa$ is an appropriate kernel function defining the covariance structure. Given initial datasets $\mathcal{D}_{low} = \{\bm{X}_{low}, \bm{y}_{low}\}$ and $\mathcal{D}_{high} = \{\bm{X}_{high}, \bm{y}_{high}\}$, the mean and variance prediction of the low-fidelity GP is,

\begin{equation}\label{eqn:mfgprlf}
\begin{aligned}
    \mu_{low}(\bm{x}) & = \kappa_{low}(\bm{x}, \bm{X}_{low})\bm{K}_{low}^{-1}(\bm{y}_{low})\\
    \sigma_{low}^2(\bm{x}) & = \kappa_{low}(\bm{x}, \bm{x}) -\kappa_{low}(\bm{x}, \bm{X}_{low})\bm{K}_{low}^{-1}\kappa_{low}( \bm{X}_{low}, \bm{x}).\\
\end{aligned}
\end{equation}

\noindent The mean of the low-fidelity GP is transferred to the high-fidelity GP, including scaling by the hyperparameter $\rho$. The correction GP $\bm{y}$ values constitute the difference between the high-fidelity function values and the mean predictions of the scaled low-fidelity GP. The variance of a scaled GP is affected by the square of the scaling factor. Thus, the uncertainty of the low-fidelity model is transferred to the high-fidelity model, but multiplied by  $\rho^2$. To construct the {\em correction GP} $\delta(\bm{x})$, the high-fidelity sampling points should ideally be a subset of the low-fidelity sampling points. Nested sampling techniques \cite{nestedlhs} can be used to initialize such datasets. This correction GP accounts for the uncertainty in the mismatch between the low-fidelity and high-fidelity models. The mean prediction and variance of the high-fidelity GP are

\begin{equation}\label{eqn:mfgprhf}
\begin{aligned}
    \mu_{high}(\bm{x}) & = \rho \cdot \mu_{low}(\bm{x})
    + \kappa_{\delta}(\bm{x}, \bm{X}_{high})\bm{K}_{\delta}^{-1}(\bm{y}_{high} - \rho \cdot \mu_{low}(\bm{X}_{high}))\\
    \sigma_{high}^2(\bm{x}) & = \rho^2 \cdot \sigma_{low}^2(\bm{x})
    + \kappa_{\delta}(\bm{x}, \bm{x}) -\kappa_{\delta}(\bm{x}, \bm{X}_{high})\bm{K}_{\delta}^{-1}\kappa_{\delta}( \bm{X}_{high}, \bm{x}).\\
\end{aligned}
\end{equation}

Although nested sampling points ($\bm{X}_{high} \subset \bm{X}_{low}$) are typically used to initialize multi-fidelity GPs, non-nested sampling points can also be used. In such cases, the discrepancy at high-fidelity locations not included in the low-fidelity locations is estimated using the mean prediction of the low-fidelity GP. 

\begin{figure}[H]
     \centering
     \begin{subfigure}[b]{0.49\textwidth}
         \centering
         \includegraphics[width=\textwidth]{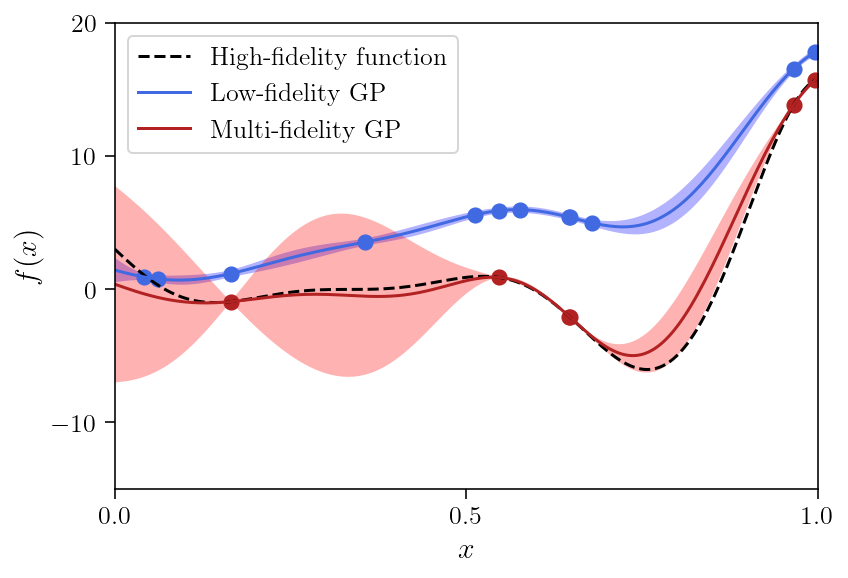}
     \end{subfigure}
     \hfill
     \begin{subfigure}[b]{0.49\textwidth}
         \centering
         \includegraphics[width=\textwidth]{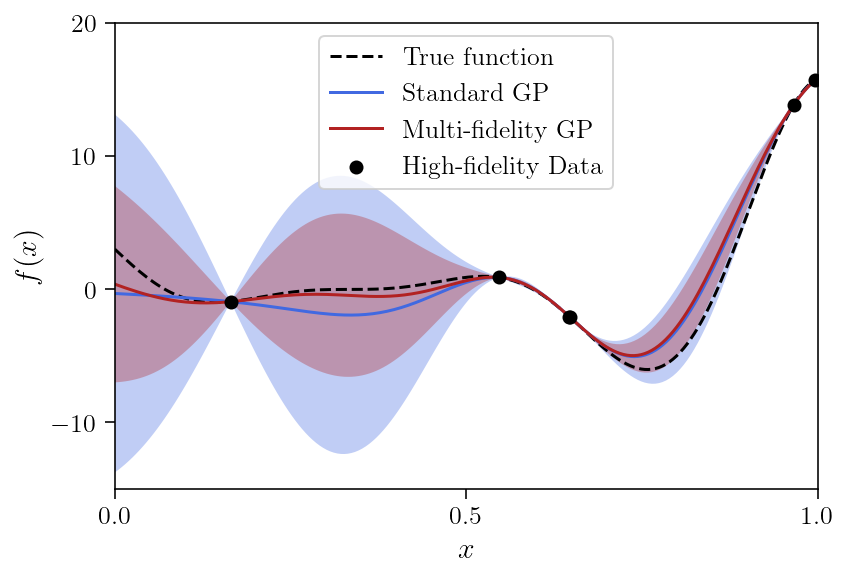}
     \end{subfigure}
     \hfill
     \caption{\textbf{Multi-fidelity Gaussian Process Regression.} (a) This illustrative figure shows the Multi-fidelity GP constructed using many low-fidelity data points (blue) and few high-fidelity data points (red), where the high-fidelity samples are a subset of the low-fidelity sampling points. (b) In comparison with a Standard GP constructed using just the high-fidelity data points, we observe that the Multi-fidelity GP obtains better mean prediction and lower variance. The shaded regions represent the 95\% confidence intervals predicted by the GPs.}
     \label{fig:mfgpvgp}
\end{figure}

\subsection{Multi-fidelity Acquisition Functions}
In this section, we will explore three distinct multi-fidelity acquisition functions. In addition to providing the next best location, these acquisition functions also select at which level of fidelity to evaluate next.
\subsubsection{Fidelity-Weighted Acquisition Function}
For both low-fidelity and high-fidelity GPs, we use a base acquisition function (e.g., UCB \cite{UCb}, EI \cite{ei}) which is adjusted using a cost-ratio penalty \cite{sapsis} term to account for the significant difference in computational expense between the high-fidelity and low-fidelity function. Both of these modified acquisition functions are optimized, and the best value between the two decides the location and fidelity for the next evaluation. By appropriately adjusting this cost-ratio penalty, we bias the fidelity-weighted acquisition function towards selecting points from the low-fidelity function more frequently. The cost-ratio penalty term can be expressed as,
\begin{equation}\label{cost}
    C_{low} = \frac{\lambda_1}{\lambda_2}(n_1 + 1) + n_2
    \quad\text{and}\quad
    C_{high} = \frac{\lambda_1}{\lambda_2}n_1 + (n_2+1).
\end{equation}

\noindent The parameters $\lambda_1$ and $\lambda_2$ are tunable cost parameters that control sampling from the low-fidelity and high-fidelity models, respectively. $n_1$ and $n_2$ represent the number of evaluations of the low-fidelity and high-fidelity models, respectively. Assuming that we are maximizing the acquisition function, the negative of the cost-ratio penalty is added to the base acquisition function to obtain the Fidelity-Weighted Acquisition Function. Maximizing this term balances maximizing the base acquisition function while minimizing the cost-ratio penalty. 
\begin{equation}\label{acq}
    \alpha_{low}^{fw}(\bm{x}) = \alpha_{low}(\bm{x}) - \frac{1}{n_{iter}}C_{low}
    \quad\quad
    \alpha_{high}^{fw}(\bm{x}) = \alpha_{high}(\bm{x}) - \frac{1}{n_{iter}}C_{high}.
\end{equation}
 The {\em next evaluation point} ($\bm{x}_{t}$) and {\em the corresponding fidelity} ($s_{t}$) are determined by independently optimizing the acquisition functions at each fidelity level and selecting the point with the highest overall acquisition value.

\begin{equation}
    \bm{x}_{t} = \argmax_{\bm{x} \in \mathcal{X}}
    (\max(\alpha_{low}^{fw}(\bm{x}), \alpha_{high}^{fw}(\bm{x})))
\end{equation}
\begin{equation}
    s_{t} = \argmax_{s \in \{low, high\}}(\alpha_{s}^{fw}(\bm{x}))
\end{equation}

\noindent The cost parameters should be tuned in a way that both the cost-ratio penalty term {\em and the base acquisition function} are taken into account while optimizing the Fidelity-Weighted Acquisition Function.

\begin{algorithm}[H]
\caption{Fidelity-Weighted Strategy}
\begin{algorithmic}
\Require Functions $f_{\text{low}}(\bm{x})$ and $f_{\text{high}}(\bm{x})$; Initial datasets $\mathcal{D}_{\text{low}}, \mathcal{D}_{\text{high}}$
\For{$t = 1$ to $T$}
    \State Train $\mathcal{MFGP}$ using $\mathcal{D}_{low}$ and $\mathcal{D}_{high}$
    \State $\bm{x}_{t} \in \arg\max_{\bm{x} \in \mathcal{X}}(\alpha_{low}^{fw}(\bm{x}), \alpha_{high}^{fw}(\bm{x}))$ \Comment{Next location}
    \State $s_{t} \in \argmax_{s \in \{low, high\}}(\alpha_{s}^{fw}(\bm{x}))$ \Comment{Next fidelity}
    \State $\bm{y}_{t} \gets f_{s_t}(\bm{x}_t)$ \Comment{Function evaluation}
    \State $\mathcal{D}_{s_t} \gets \mathcal{D}_{s_t} \cup \{(\bm{x}_t, \bm{y}_t)\}$ 
\EndFor
\State $\bm{x}^* \gets \arg\max_{\bm{x} \in \mathcal{D}_{high}} f_{high}(\bm{x})$ \Comment{Best high-fidelity solution}
\end{algorithmic}
\end{algorithm}

\subsubsection{Multi-fidelity UCB}
Kandasamy et al.\cite{mfucb} proposed a separate multi-fidelity acquisition function based on upper confidence bounds, and observed improved/useful information exchange between fidelities. The acquisition functions of both their low-fidelity and high-fidelity GPs were based on the Upper Confidence Bound \cite{UCb}, which balances the exploration of uncertain regions ($\sigma$) with the exploitation of known ``good" regions ($\mu$) using a hyperparameter $ \beta $. However, an additional error-term ($\zeta$) is added to the low-fidelity UCB: this error term encodes an (assumed available) error bound between the high-fidelity and low-fidelity objective function. This widens the confidence bound of the low-fidelity UCB:
\begin{equation}
    \alpha_{low}(\bm{x}) = \mu_{low}(\bm{x}) + \beta^{1/2}\sigma_{low}(\bm{x}) +\zeta
\end{equation}

\begin{equation}
    \zeta(\bm{x}) = \norm{f_{high}(\bm{x}) - f_{low}(\bm{x})}
\end{equation}

\begin{equation}
    \alpha_{high}(\bm{x}) = \mu_{high}(\bm{x}) + \beta^{1/2}\sigma_{high}(\bm{x}). 
\end{equation}

\begin{figure}[H]
     \centering
     \begin{subfigure}[b]{0.49\textwidth}
         \centering
         \includegraphics[width=\textwidth]{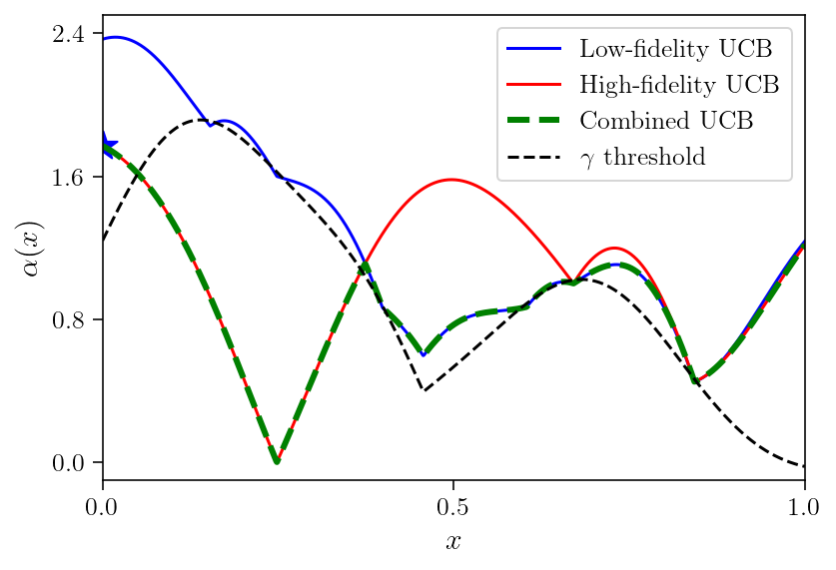}
         
     \end{subfigure}
     \hfill
     \begin{subfigure}[b]{0.49\textwidth}
         \centering
         \includegraphics[width=\textwidth]{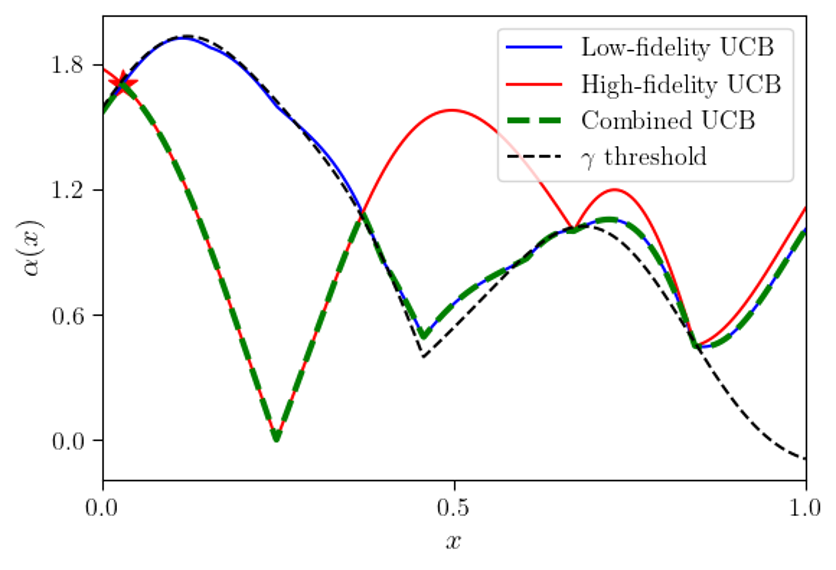}
     \end{subfigure}
     \hfill
     \caption{\textbf{Multi-fidelity Upper Confidence Bound}. Demonstration of the $\gamma$ bound criterion: High-fidelity and low-fidelity UCB are marked red and blue respectively. The combined best upper bound is denoted by the green dashed line. The black dotted line denotes the $\gamma$ threshold. The red and blue stars denote the next best location obtained by maximizing the combined best bound. (a) In this plot, we see an example when the low-fidelity variance term is higher than $\gamma$. In this iteration, the algorithm will select to perform a low-fidelity function evaluation next. (b) When the low-fidelity variance term drops below $\gamma$, the algorithm selects to perform a high-fidelity evaluation next in this location}
     \label{fig:gamma_threshold}
\end{figure}

\noindent Given these two upper confidence bounds, the combined best upper bound for the high-fidelity function is found by taking the minimum of both, 

\begin{equation}
    \bm{x}_{t} = \argmax_{\bm{x} \in \mathcal{X}}(min(\alpha_{low}(\bm{x}), \alpha_{high}(\bm{x}))).
\end{equation}
Once we obtain the next function evaluation location, the next fidelity level to employ for the function evaluation is selected by a threshold condition $\gamma$. 
As illustrated in Figure~\ref{fig:gamma_threshold}, if the variance term in the low-fidelity aquisition function is high (because we have not yet adequately sampled the low-fidelity function in this region) then an additional low-fidelity evaluation is performed; else, if the low-fidelity variance term is lower than the threshold, then we consider we have sufficient information from the low-fidelity model to go ahead with performing a high-fidelity evaluation. 
The threshold value ($\gamma$) is defined by the tunable cost parameters ($\lambda_1$ and $\lambda_2$) and the error bound ($\zeta$). For practical problems in which we do not have good estimates of the difference between low- and high-fidelity objective functions (i.e. of the error bound, $\zeta$), this error is estimated pointwise by taking the absolute difference between the GP means. 
As the optimization algorithm progresses, this value improves, and will hopefully estimate the error more accurately.

\begin{algorithm}[H]
\caption{MF-GPR-UCB Optimization}
\begin{algorithmic}
\Require Functions $f_{\text{low}}(\bm{x})$ and $f_{\text{high}}(\bm{x})$; Initial datasets $\mathcal{D}_{\text{low}}, \mathcal{D}_{\text{high}}$
\For{$t = 1$ to $T$}
    \State Train $\mathcal{MFGP}$ using $\mathcal{D}_{low}$ and $\mathcal{D}_{high}$
    \State $\bm{x}_t \in \arg\max_{\bm{x} \in \mathcal{X}} \min(\alpha_{\text{low}}(\bm{x}), \alpha_{\text{high}}(\bm{x}))$ \Comment{Next location}
    \State $\zeta \gets \norm{\mu_1(\bm{x}_t) - \mu_2(\bm{x}_t)}$
    \State $\gamma \gets \zeta \sqrt{\lambda_2 / \lambda_1}$ \Comment{Threshold criterion}
    \If{$\beta^{1/2} \sigma_{\text{low}}(\bm{x}_t) > \gamma$} 
        \State $s_t \gets \text{low}$ 
    \Else
        \State $s_t \gets \text{high}$ 
    \EndIf \Comment{Next fidelity}
    \State $\bm{y}_t \gets f_{s_t}(\bm{x}_t)$ \Comment{Function evaluation}
    \State $\mathcal{D}_{s_t} \gets \mathcal{D}_{s_t} \cup \{(\bm{x}_t, \bm{y}_t)\}$ 
\EndFor
\State $\bm{x}^* \gets \arg\max_{\bm{x} \in \mathcal{D}_{\text{high}}} f_{\text{high}}(\bm{x})$ \Comment{Best high-fidelity solution}
\end{algorithmic}
\end{algorithm}

\subsubsection{Proximity-based Acquisition Function}
In this approach, instead of using two acquisition functions, we only utilize the acquisition function of the high-fidelity surrogate model ($\alpha_{high}$). This ensures that the next evaluation point is selected based on the most accurate representation of the objective function. However, rather than defaulting to a high-fidelity evaluation {\em at each iteration}, which may be computationally expensive, we introduce a decision-making criterion that considers {\em the local density of low-fidelity data}.

\begin{figure}[H]
    \centering
    \includegraphics[width=0.8\linewidth]{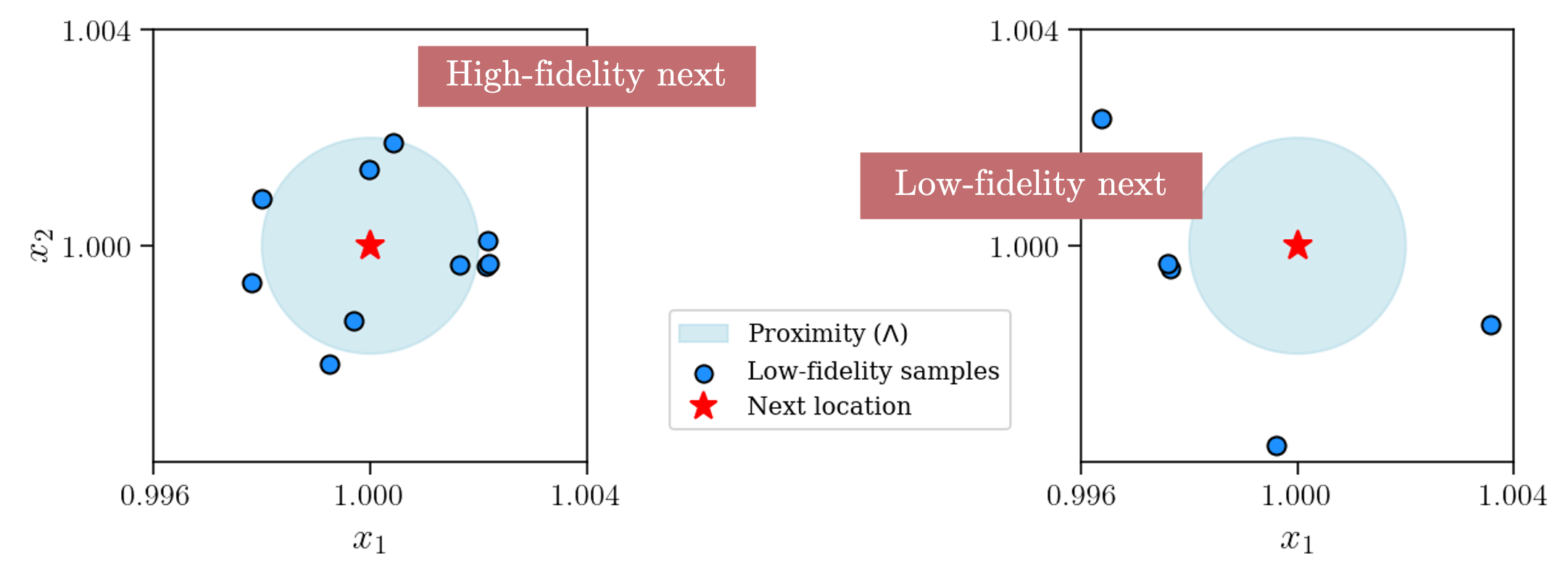}
    \caption{\textbf{Proximity-based acquisition function.} The proximity strategy is demonstrated for a two-dimensional parameter space to illustrate fidelity selection. Depending on the local density of low-fidelity data points, the acquisition performs fidelity selection. The proximity region is defined using the cost ratio ($\Lambda = \lambda_1/\lambda_2$)}
    \label{fig:prox_acq}
\end{figure}

Specifically, once the next-best location is determined by maximizing the high-fidelity acquisition function, we assess whether nearby low-fidelity observations are already available within a specified neighborhood of that point. If such low-fidelity data are not sufficiently dense in the vicinity, we may defer the high-fidelity evaluation in favor of a lower-cost, low-fidelity query. In contrast, if the region has adequate low-fidelity coverage, a high-fidelity evaluation should be performed. The proximity parameter is defined using the tunable cost parameters ($\lambda_1$ and $\lambda_2$), which reflect the relative costs of acquiring low- and high-fidelity data, respectively. As this cost difference increases, the cost-ratio, and importantly, the radius of the proximity region (Figure \ref{fig:prox_acq}), shrinks, encouraging more low-fidelity evaluations before reverting back to high-fidelity.

\begin{algorithm}[H]
\caption{Proximity-based Strategy}
\begin{algorithmic}
\Require Functions $f_{\text{low}}(\bm{x})$ and $f_{\text{high}}(\bm{x})$; initial datasets $\mathcal{D}_{\text{low}}, \mathcal{D}_{\text{high}}$
\For{$t = 1$ to $T$}
    \State Train $\mathcal{MFGP}$ using $\mathcal{D}_{low}$ and $\mathcal{D}_{high}$
    \State $\bm{x}_t \gets \arg\max_{\bm{x} \in \mathcal{X}} \alpha_{\text{high}}(\bm{x})$ \Comment{Next location}
    \If{$\|\bm{x}_t - \bm{x}_{\text{low}}\| > \lambda_1 / \lambda_2$}
    \Comment{Proximity criterion}
        \State $s_t \gets \text{low}$ 
    \Else
        \State $s_t \gets \text{high}$ 
    \EndIf \Comment{Next fidelity}
    \State $\bm{y}_t \gets f_{s_t}(\bm{x}_t)$ \Comment{Function evaluation}
    \State $\mathcal{D}_{s_t} \gets \mathcal{D}_{s_t} \cup \{(\bm{x}_t, \bm{y}_t)\}$
\EndFor
\State $\bm{x}^* \gets \arg\max_{\bm{x} \in \mathcal{D}_{\text{high}}} f_{\text{high}}(\bm{x})$ \Comment{Best high-fidelity solution}
\end{algorithmic}
\end{algorithm}

\subsection{Test functions} 
\begin{figure}[H]
    \centering
    \begin{subfigure}[b]{0.49\textwidth}
         \centering
         \includegraphics[width=\textwidth]{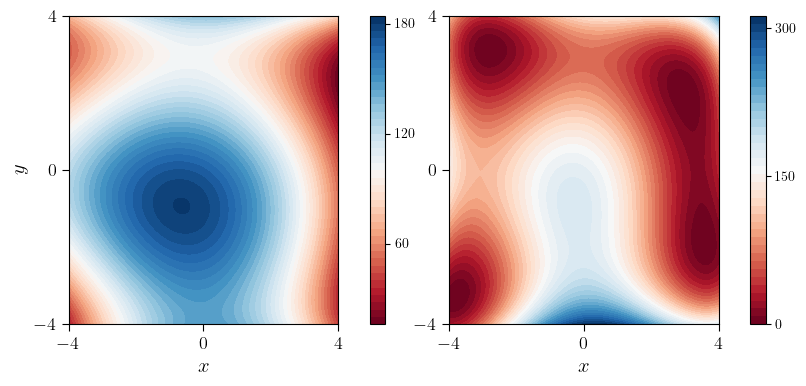}
     \end{subfigure}
     \hfill
     \begin{subfigure}[b]{0.49\textwidth}
         \centering
         \includegraphics[width=\textwidth]{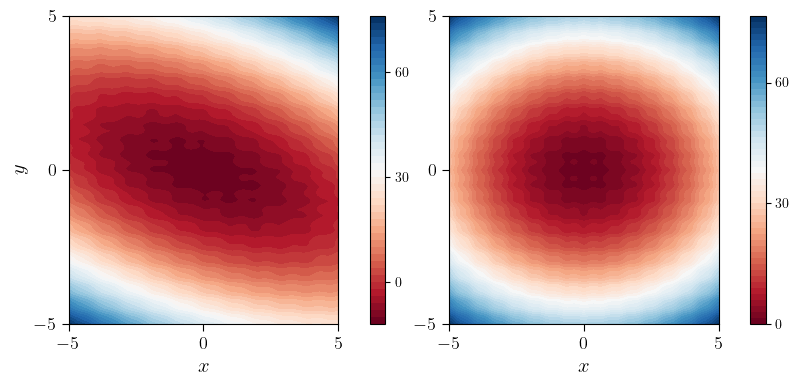}
     \end{subfigure}
     \hfill
     \caption{\textbf{Synthetic test functions.} (a) Low-fidelity and high-fidelity Himmelblau function \cite{bohachevskyhimmelblau}. (b) Low-fidelity and high-fidelity Bohachevsky function.}
     \label{fig:test2d}
  \end{figure}

We test the performance of the acquisition functions with multi-fidelity GPs across various problems, starting with synthetic test functions like the Himmelblau function (Figure \ref{fig:test2d}), which has four high-fidelity global optima out of which only one is located near the low-fidelity optimum, and the Forrester function, where the low-fidelity optimum lies close to a high-fidelity local optimum, far away from the true global optimum. In addition, we benchmark the multi-fidelity framework on problems including (a) an enzyme reaction scheme --as well as a nonlinear reaction model--  where the application of the Quasi Steady-State Approximation \cite{rawlings2002chemical} creates an (imperfect) low-fidelity model; and (b) a periodically-forced ammonia catalysis dynamic model where varying time-integration tolerances creates two distinct fidelity models. More information on all the test functions presented here can be found in \ref{app:tf}.

\subsubsection{Toy Enzyme Model}
\noindent Evangelou et al. \cite{nikoenzyme} studied a simple enzymatic reaction scheme \cite{enzymeqssa} consisting of four ODEs (\ref{eqn:hfenzyme}). This serves as the high-fidelity model. 
\begin{equation}\label{eqn:enzyme}
    E + S_0  \underset{k_r}{\stackrel{k_f}{\rightleftharpoons}} ES_0 \xrightarrow{k_{\text{cat}}} E + S_1
\end{equation}
\begin{equation}\label{eqn:hfenzyme}
\begin{aligned}
    \frac{d[S_0]}{dt} &= -k_f[E][S_0] + k_r[ES_1] \\
    \frac{d[ES_0]}{dt} &= k_f[E][S_0] - k_r[ES_0] - k_{\text{cat}}[ES_0] \\
    \frac{d[S_1]}{dt} &= k_{\text{cat}}[ES_0] \\
    \frac{d[E]}{dt} &= -k_f[E][S_0] + k_r[ES_0] + k_{\text{cat}}[ES_0]
\end{aligned}
\end{equation}

\noindent The low-fidelity model is a Quasi Steady-State Approximation (QSSA) of the enzyme model consisting now of two ODEs. Evangelou et al., used QSSA for the species $ES_0$ to approximately reduce the system. If the assumption

\begin{equation}
    S_{tot}\ll \frac{k_r + k_{cat}}{k_f} ,
\end{equation}
where
\begin{equation}
    S_{tot} = [S_0] + [S_1] + [ES_0],
\end{equation}

\noindent reasonably approximately holds, then the initial model reduces to a two-state kinetic model (\ref{eqn:lfenzyme}) with a single effective parameter which is a combination of the exact model parameters: 

\begin{equation}\label{eqn:lfenzyme}
\begin{aligned}
    \frac{d[S_0]}{dt} &= -k_{\text{eff}}[E][S_0] \\
    \frac{d[S_1]}{dt} &= k_{\text{eff}}[E][S_0] \\
    k_{\text{eff}} &= E_{\text{tot}}\frac{k_fk_{\text{cat}}}{k_r + k_{\text{cat}}}. \\
\end{aligned}
\end{equation}

\noindent Operating the QSSA model in parameter regimes where the assumption is no longer valid, results in a (initially slightly) inaccurate, low-fidelity model. To create a toy optimization problem from this model, we locate through optimization the enzyme concentration such that the reaction conversion ($X$) at a fixed given time (say 10 seconds) is 67\%. This effectively solves an algebraic equation by minimizing its residual.

\begin{equation}\label{eqn:enymeopt}
    \arg \min_E |X(E;T=10) - 0.67|
\end{equation}

\subsubsection{The Oregonator scheme: Active search for a Hopf bifurcation}
\begin{figure}[H]
    \centering
    \includegraphics[width=0.7\linewidth]{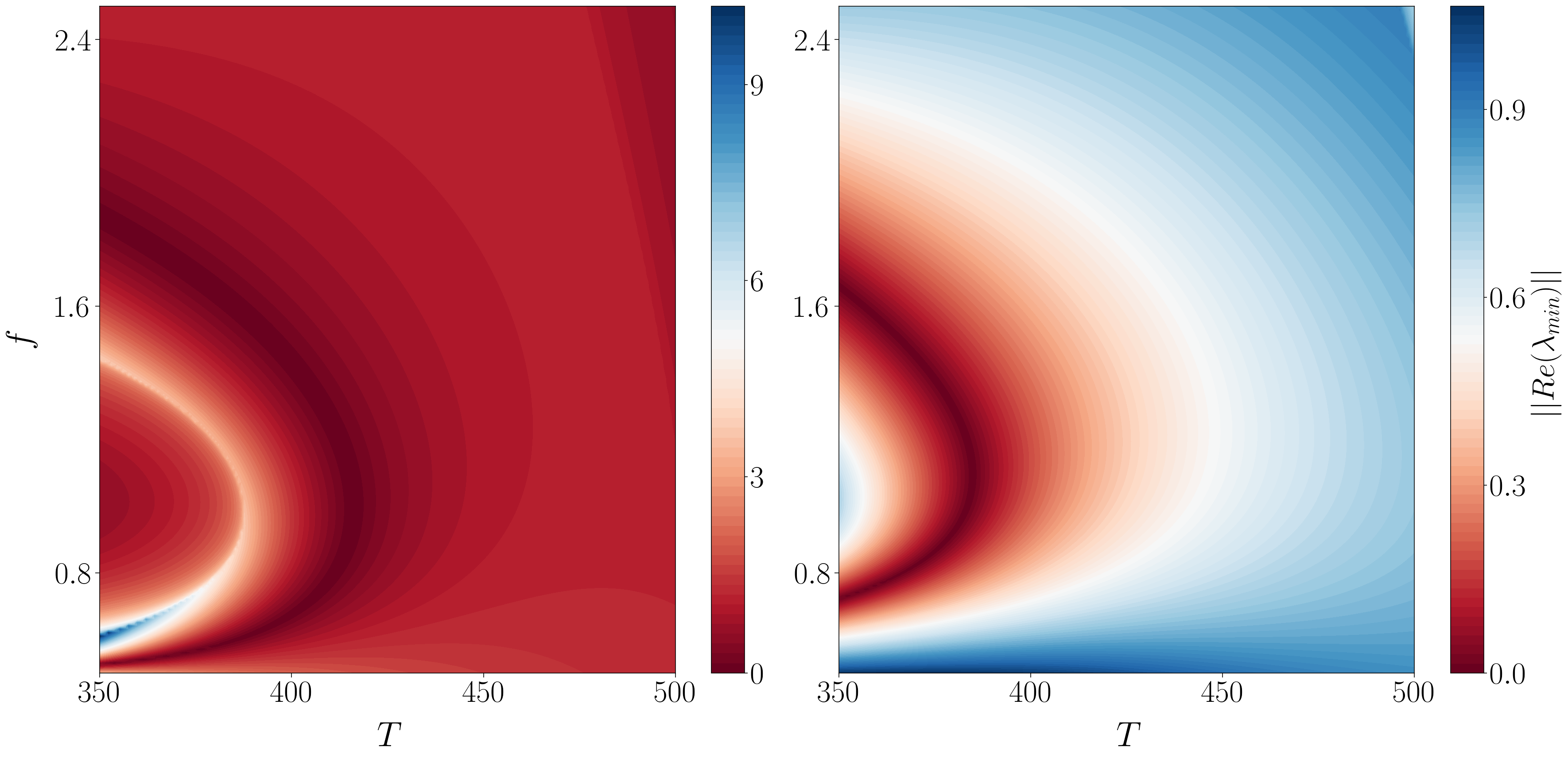}
    \caption{\textbf{Multi-fidelity Oregonator Model}. A multifidelity rendering of a two-parameter bifurcation diagram (finding the locus of Hopf bifurcation points). (a) Low-fidelity QSSA model. (b) High-fidelity exact model. The bifurcation diagrams were constructed by varying the stoichiometric factor ($f$) and the temperature ($T$). The darkest red region in both marks our multifidelity estimates of the Hopf bifurcation locus in two-parameter space. 
    The color bar indicates our estimate of the absolute value of the real part of the critical eigenvalue pair associated with the Hopf bifurcation (which occurs when the real part crosses zero). 
    The region enclosed by the convex curve exhibits limit cycle behavior.}
    \label{fig:oregonator}
\end{figure}

The Belousov-Zhabotinsky reaction is a very well studied nonlinear chemical kinetic model \cite{BZ-FKN, BZ-FKN2}. The Oregonator is a simplified representation of the system which retains key nonlinear features leading to oscillatory behavior. The onset of oscillations in such systems is marked by a Hopf bifurcation\cite{angeli2004bifurcdetection, kevrekidishopf, psarellis2025active, marsden1976hopf};
at a supercritical Hopf bifurcation point, varying an operating parameter past a critical value shifts the attractor of a dynamical system from a stable fixed point that loses stability to a stable periodic orbit. Pullela et al. \cite{oregonator(T)} explored the dynamics of an Oregonator model over a wide range of temperature values, initial reagent concentration ratio ($b/a$), and stoichiometric factors ($f$). More information on the temperature-dependent oregonator model can be found in \ref{app:tf}. A multi-fidelity model (Figure \ref{fig:oregonator}) can be constructed by applying QSSA to reduce the 3D oregonator model into a two-dimensional model. The full oregonator model is,

\begin{equation}\label{eqn:oreg_full}
\begin{aligned}
    \varepsilon(T)\dot x &= q(T)ay - xy + ax -x^2\\
    \omega(T)\dot y &= -q(T)ay - xy +fbz\\
    \dot z &= ax - bz.
\end{aligned}
\end{equation}

\noindent Using the QSSA approximation $\varepsilon(T) \ll \omega(T)$, we can reduce the model into a pair of differential equations,

\begin{equation}\label{eqn:oreg_red}
\begin{aligned}
    \omega(T)\dot y &= -q(T)ay - x^*y +fbz\\
    \dot z &= ax^* - bz \\
    x^* &= \frac{a - y}{2} + \sqrt{q(T)ay + \frac{(a-y)^2}{4}}.
\end{aligned}
\end{equation}

\noindent We can detect the Hopf bifurcation by simultaneously solving for a steady-state and a criticality condition. Fixing $b/a$, the critical parameter $p^* = (T^*, f^*)$ can be identified 
by reformulating the bifurcation detection as an optimization problem, trying to ``push"  the real part of the critical eigenvalue pair of the system linearization to zero,  
\begin{equation}\label{oreg_opt}
    p^* = arg\min_p(\norm{Re(\lambda(x^*,p))}).
\end{equation}
\noindent Here $\lambda$ is the eigenvalue of the Jacobian $J(x^*, p)$ at steady-state solution $x^*$. Typically, it is necessary to test whether higher-order derivative-based {\em nondegeneracy conditions} are satisfied; we do not consider this feature here.

\subsubsection{Dynamic Ammonia Catalysis: Optimal waveforms}
\begin{figure}[H]
     \centering
     \begin{subfigure}[b]{0.64\textwidth}
         \centering
         \includegraphics[width=\textwidth]{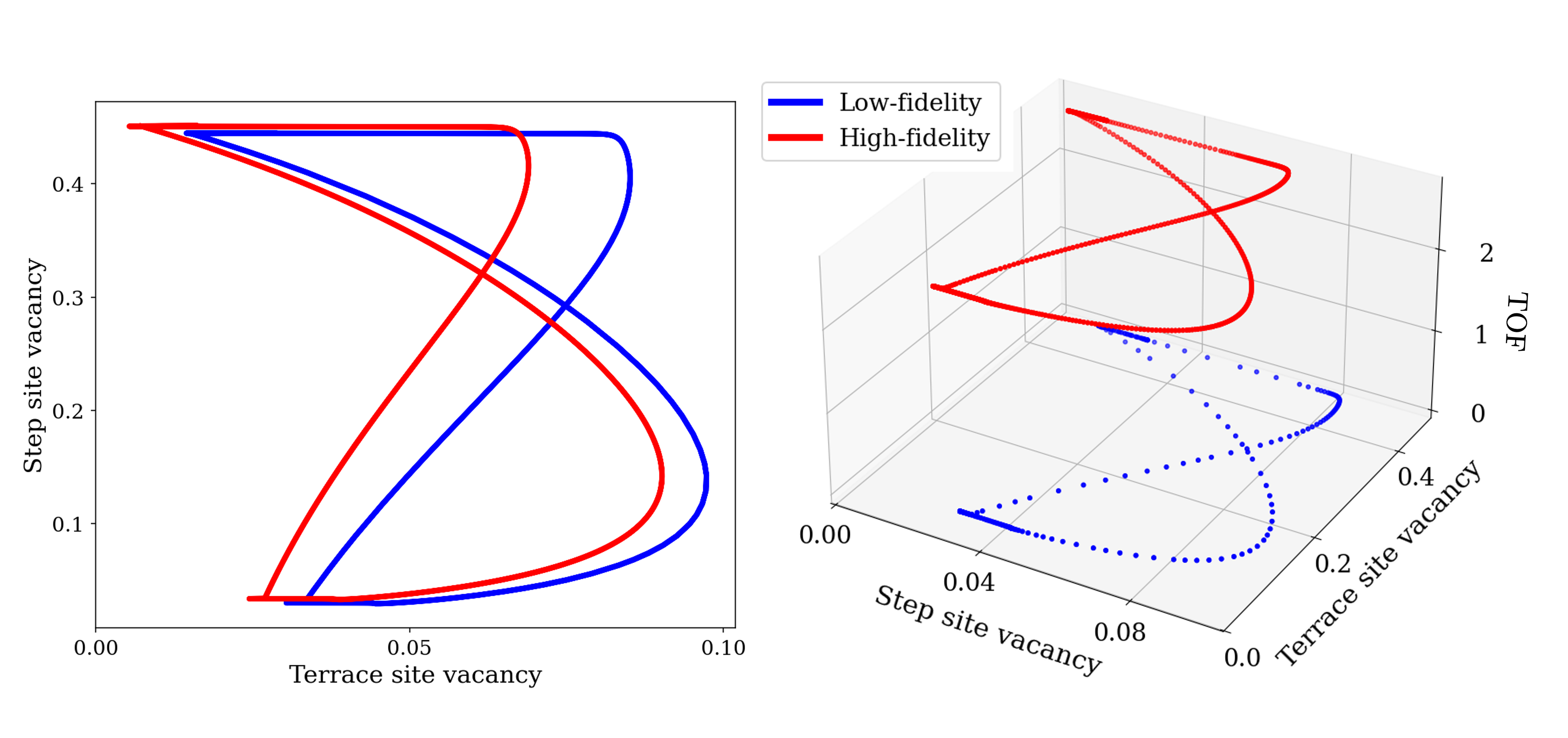}
     \end{subfigure}
     \hfill
     \begin{subfigure}[b]{0.33\textwidth}
         \centering
         \includegraphics[width=\textwidth]{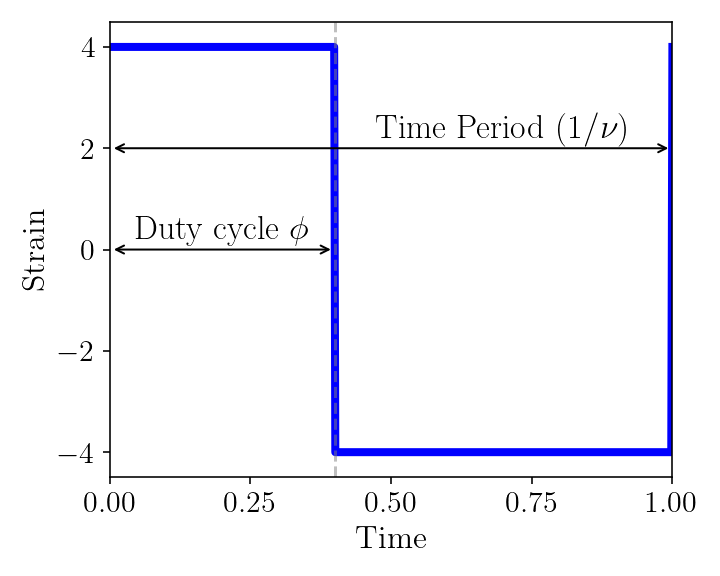}
     \end{subfigure}
     \hfill
     \caption{\textbf{Multi-fidelity Ammonia Catalysis Model}. (a) {\em Periodically forced limit cycles}. Left: 2D phase portrait of the multi-fidelity limit cycles.  Right: 3D phase portrait of the multi-fidelity limit cycles. The catalyst site vacancies are plotted at the high-fidelity periodic steady-state and the low-fidelity approximation. The 3D plot demonstrates the difference in TOF between the levels of fidelity. The low-fidelity model was created by varying the relative time-integration tolerance from $10^{-8}$ to $10^{-4}$.  (b) {\em Dynamic Strain wave}. Variation of the catalyst surface strain \cite{dyncatalysis, ammonia} between two states using a square wave parametrized using the oscillation frequency ($\nu$) and duty cycle ($\phi$).}
     \label{fig:ammonia}
\end{figure}

    Wittreich et al. developed a 16-chemical species ammonia catalysis model to investigate strain effects on a Ruthenium catalyst with step and terrace sites \cite{ammonia}. Density Functional Theory (DFT) calculations revealed that varying the strain alters the adsorbate binding energies of the catalyst\cite{dyncatalysis}. The objective is to enhance catalyst performance by oscillating the applied strain, particularly under low pressure conditions \cite{ammonia, programmablecatalyst}. The turnover frequency (TOF) is an excellent indicator of catalyst performance and is typically defined as the number of reaction product molecules generated per active catalyst site per unit time. This optimization problem focuses on maximizing the TOF of the catalyst as a function of the strain forcing waveform parameters. We oscillate the strain between +4\% strain and -4\% strain using a square wave (Figure \ref{fig:ammonia}) characterized by the strain oscillation frequency ($\nu$) and the duty cycle ($\phi$). The duty cycle of the square wave describes the fraction of time it spends in its ``high" state during one full period. A low-fidelity model can be created by either developing a reduced kinetic model or by using less stringent time-integration tolerances. In this work, we perform a brute-force integration for up to a thousand cycles on a low-accuracy model which provides an inaccurate solution ($x_{low}^*$), often far from the actual eventual ``periodic steady-state". The low-fidelity model is a period-averaged TOF calculated on this final low-fidelity state. The high-fidelity model calculates the period-averaged TOF in the actual periodic steady-state ($x_{high}^*$) through a matrix-free algorithm for solving boundary value problems in time with Newton-Krylov GMRES \cite{kelley1995iterative, kelley2004newton} iterations:

\begin{equation}\label{eqn:ammonia_model}
\begin{aligned}
    \nu^*, \phi^* &= arg \max_{\nu, \phi} (\hat {TOF}(\nu, \phi)).
\end{aligned}
\end{equation}

\begin{table}[H]
    \centering
    \begin{tabular}{l|c|c|c|c}
        \toprule
        \textbf{Test function} & \textbf{Dimensionality} & \textbf{Low-fidelity samples} & \textbf{High-fidelity samples}\\ 
        \midrule
        Forrester & 1 & 4 & 1 \\ 
        Toy Enzyme & 1 & 4 & 1 \\ 
        Bohachevsky & 2 & 12 & 3 \\
        Himmelblau & 2 & 12 & 3 \\
        Oregonator & 2 & 12 & 3 \\
        Ammonia & 2 & 12 & 3 \\
        \bottomrule
    \end{tabular}
    
    \caption{\textbf{Test functions.} Summary of all the test functions used. }
    \label{tab:test_funcs}
\end{table}

\section{Results and Discussion}

\subsection{Optimization Performance}

\noindent To test the performance of these algorithms, we repeat the optimization run for up to 50 different initializations. This is carried out for two explorative modes ($\beta=0.5$ and $\beta=1$) and two exploitative modes ($\beta=3$ and $\beta=5$) along with an adaptive exploration approach by varying the parameter $\beta$. We also assessed the impact of the cost parameters by repeating the optimization runs for multiple cost ratio ($\Lambda = \lambda_1/\lambda_2$) values.
For our one-dimensional illustrative functions --we include two of these in our example problem list (see Table \ref{tab:test_funcs}), each optimization run is initialized using a sparse dataset comprised of four low-fidelity samples and one high-fidelity sample. For our two-dimensional function examples, this sparse initial dataset contains 12 low-fidelity samples and three high-fidelity samples. The performance evaluation focuses {\em solely on high-fidelity solutions}, as optimizing the low-fidelity function is not the primary objective, and achieving its optimum provides no direct benefit. We use expected improvement (EI) \cite{ei} as the base acquisition function for the fidelity-weighted method and the proximity-based method. To implement varying levels of exploration and exploitation within EI, we adopt the weighted EI formulation ~\cite{sobester2005weightedEI}, introducing a tunable parameter $\beta$, similar in spirit to the trade-off control in UCB methods. More information on these acquisition functions can be found in \ref{app:acq}.

For optimization problems constrained by a total evaluation budget, it is also recommended to reserve a few (or a single) high-fidelity evaluations for the final stage \cite{mfreactor}. These evaluations can be performed exploitatively by targeting the location with the best high-fidelity GP mean. If the high-fidelity GP-mean points to a location different from the best solution found, we do perform this additional high-fidelity evaluation.

\begin{figure}[H]
     \centering
     \begin{subfigure}[b]{0.32\textwidth}
         \centering
         \includegraphics[width=\textwidth]{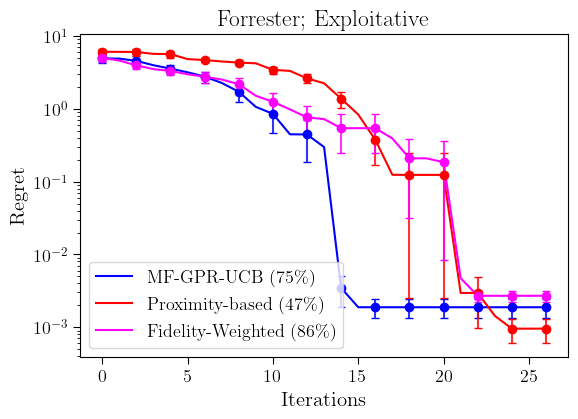}
     \end{subfigure}
     \hfill
     \begin{subfigure}[b]{0.32\textwidth}
         \centering
         \includegraphics[width=\textwidth]{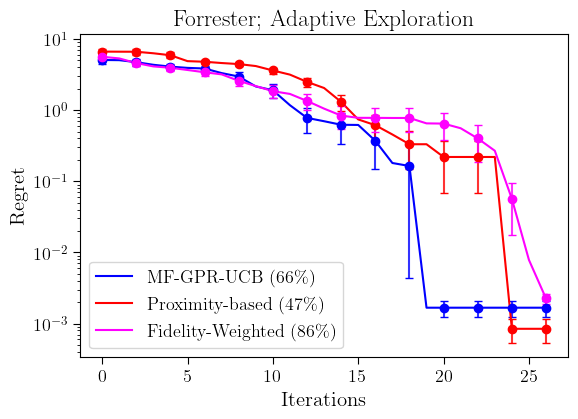}
     \end{subfigure}
     \hfill
     \begin{subfigure}[b]{0.32\textwidth}
         \centering
         \includegraphics[width=\textwidth]{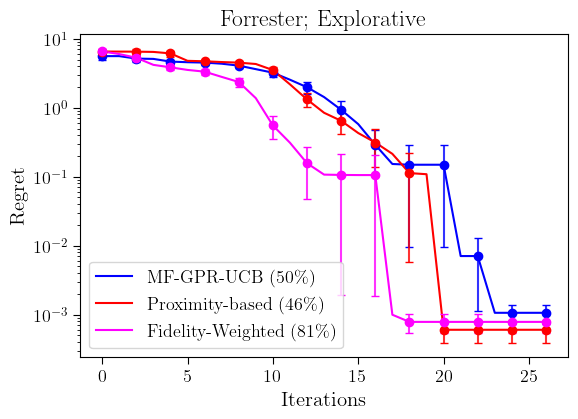}
     \end{subfigure}
     
     \caption{\textbf{Forrester 1D.} (a) Exploitative ($\beta=0.5$) (b) Adaptive exploration (c) Explorative ($\beta = 3$). All the plots display the best high-fidelity solution plotted against the number of iterations. The number in the parentheses denotes the percentage of high-fidelity evaluations used. The proximity-based acquisition function consistently converges to the global optimum using a lower percentage of high-fidelity evaluations. }
      \label{fig:forrester_performance}
\end{figure}

For the one-dimensional Forrester function, the high-fidelity function has a local optimum that coincides with the global optimum of the low-fidelity function. The algorithms may end up converging near the local optimum for this reason. We filter the optimization runs which found the global optimum for the plots. Table \ref{tab:forrester_performance} shows the percentage of optimization runs that found the global optimum.
The fidelity-weighted method consistently gets stuck in the local optimum due to (a) poor information exchange between the high-fidelity and low-fidelity acquisition functions, and (b) fewer high-fidelity evaluations for the lower cost-parameter runs. Unlike the other two multi-fidelity acquisition functions, the fidelity-weighted acquisition functions point to the optimum value at both fidelities. We also observed that some exploitative optimization runs resulted in the algorithm getting stuck near the low-fidelity optimum. The proximity-based acquisition function outperforms both algorithms (Figure \ref{fig:forrester_performance}) by exhibiting greater consistency in finding the global optimum for all the cost-parameter values.

\begin{table}[H]
    \centering
    \small
    \begin{tabularx}{\textwidth}{|p{3.8cm}|C{3cm}|C{2.2cm}|C{3cm}|C{2.2cm}}
        \toprule
        \textbf{Setting} & \textbf{Value of $\bm{\beta}$} & \textbf{Fidelity-Weighted} & \textbf{MF-GPR-UCB} & \textbf{Proximity-based}\\ 
        \midrule
        Exploitative & 0.5 & 26.0 & 48.0 & \textbf{68.0}\\ 
        Exploitative & 1 & 39.7 & 58.9 & \textbf{87.1}\\
        Explorative & 3 & 40.9 & 78.6 & \textbf{92.6}\\
        Explorative & 5 & 42.3 & 85.1 & \textbf{92.9} \\
        Adaptive exploration & $\sqrt{0.2d\log(2t)}$ & 32.9 & 52.3 & \textbf{79.4}\\
        \bottomrule
    \end{tabularx}
    \caption{\textbf{Algorithm performance for the global optimum.} The values in the Table denote the percentage of optimization runs that found the global optimum. The weaker performance of the fidelity-weighted algorithm can be attributed to the poor performance of the lower cost parameter optimization runs. The other two algorithms are more consistent in finding the global optimum, even when the parameter values are low. $d$ is the problem dimension and $t$ is the iteration number.}
    \label{tab:forrester_performance}
\end{table}

\begin{figure}[H]
     \centering
     \begin{subfigure}[b]{0.32\textwidth}
         \centering
         \includegraphics[width=\textwidth]{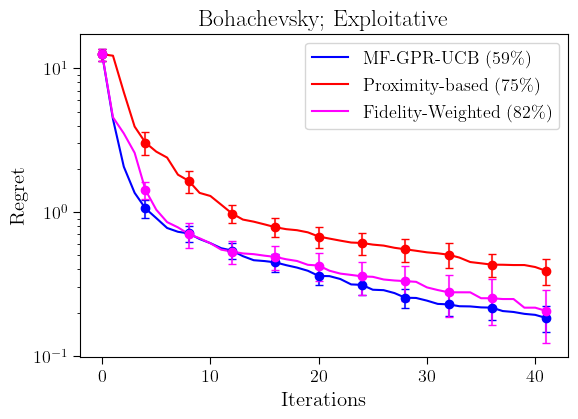}
     \end{subfigure}
     \hfill
     \begin{subfigure}[b]{0.32\textwidth}
         \centering
         \includegraphics[width=\textwidth]{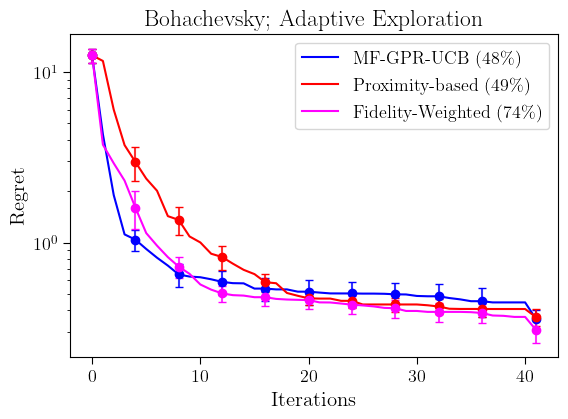}
     \end{subfigure}
     \hfill
     \begin{subfigure}[b]{0.32\textwidth}
         \centering
         \includegraphics[width=\textwidth]{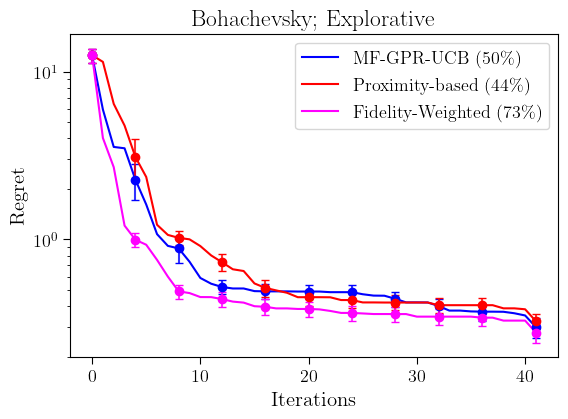}
     \end{subfigure}
     
     \caption{\textbf{Bohachevsky 2D.} (a) Exploitative ($\beta=0.5$) (b) Adaptive exploration (c) Explorative ($\beta = 3$). MF-GPR-UCB shows greater consistency over the different exploration modes, converging to the global optimum using a lower percentage of high-fidelity evaluations. Proximity-based acquisition function performs well under explorative settings with low reliance high-fidelity evaluations, but underperforms in exploitative ones. In contrast, the fidelity-weighted strategy shows a strong reliance on high-fidelity evaluations.}
      \label{fig:bohachevsky_performance}
\end{figure}

\noindent All the multi-fidelity acquisition functions have similar performance for the Bohachevsky function (Figure \ref{fig:bohachevsky_performance}). We observe that on average, the fidelity-weighted acquisition function necessitates a greater number of high-fidelity evaluations. Although the proximity-based strategy achieves a good balance between the regret achieved and the high-fidelity usage for the explorative modes, it underperforms in the exploitative regime. Among the acquisition functions, the multi-fidelity UCB exhibits the best overall performance for this test case.

\begin{figure}[H]
     \centering
     \begin{subfigure}[b]{0.32\textwidth}
         \centering
         \includegraphics[width=\textwidth]{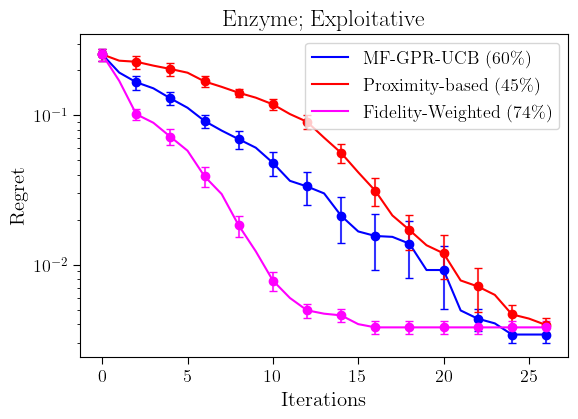}
     \end{subfigure}
     \hfill
     \begin{subfigure}[b]{0.32\textwidth}
         \centering
         \includegraphics[width=\textwidth]{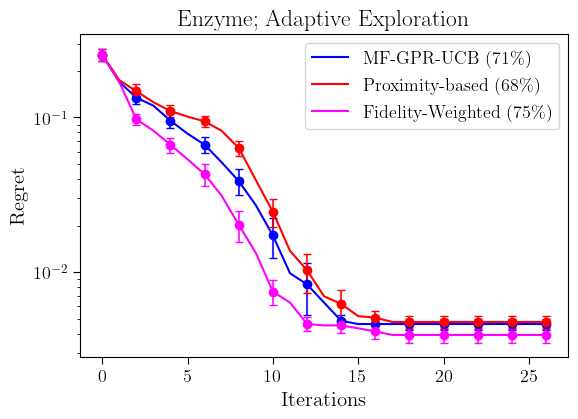}
     \end{subfigure}
     \hfill
     \begin{subfigure}[b]{0.32\textwidth}
         \centering
         \includegraphics[width=\textwidth]{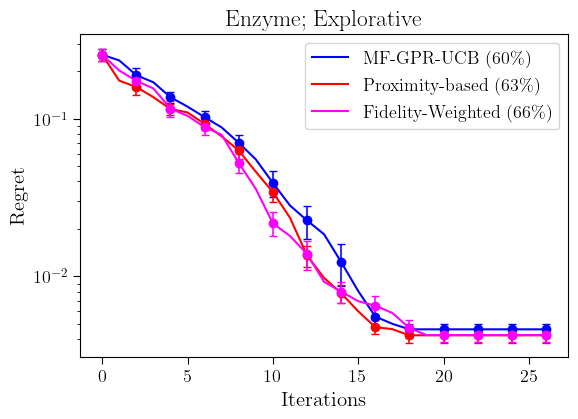}
     \end{subfigure}
     
     \caption{\textbf{Toy enzyme 1D.} (a) Exploitative ($\beta=0.5$) (b) Adaptive exploration (c) Explorative ($\beta = 3$). The proximity-based acquisition function strikes a good balance between the regret achieved and the percentage of high-fidelity evaluations
     All algorithms have similar regret performance for the exploration and adaptive exploration modes.}
      \label{fig:enzyme_performance}
\end{figure}

\noindent In the case of the toy-enzyme problem (Figure \ref{fig:enzyme_performance}), the proximity-based acquisition function achieves a better balance between the percentage of high-fidelity evaluations and the regret achieved. All three algorithms have similar performance for the explorative setting.

\begin{figure}[H]
     \centering
     \begin{subfigure}[b]{0.32\textwidth}
         \centering
         \includegraphics[width=\textwidth]{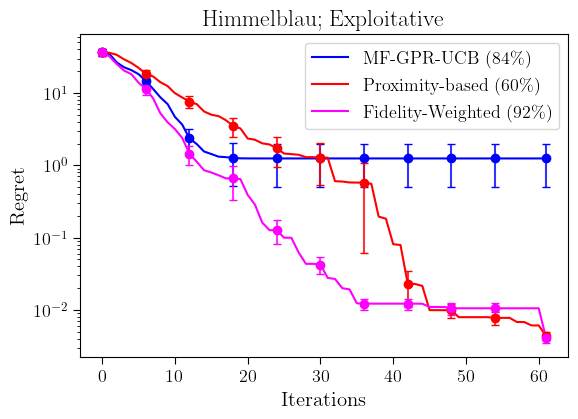}
     \end{subfigure}
     \hfill
     \begin{subfigure}[b]{0.32\textwidth}
         \centering
         \includegraphics[width=\textwidth]{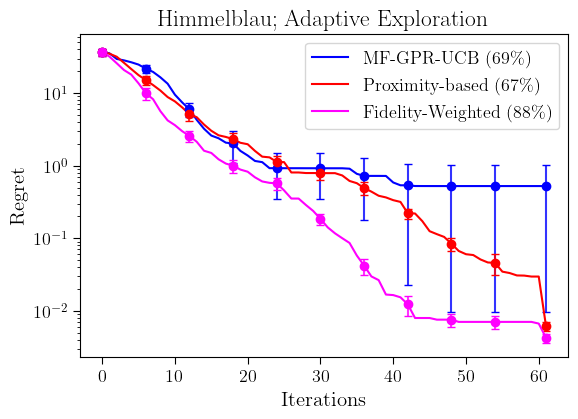}
     \end{subfigure}
     \hfill
     \begin{subfigure}[b]{0.32\textwidth}
         \centering
         \includegraphics[width=\textwidth]{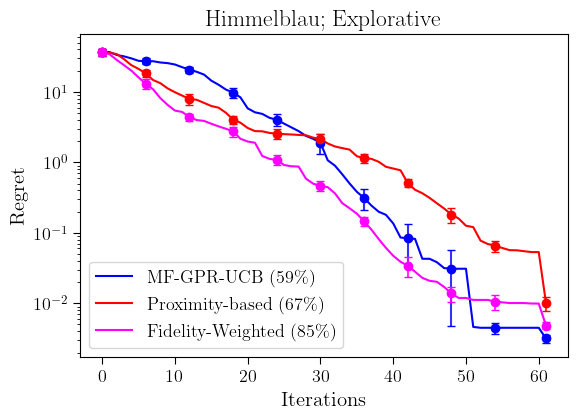}
     \end{subfigure}
     
     \caption{\textbf{Himmelblau 2D.} (a) Exploitative ($\beta=0.5$) (b) Adaptive exploration (c) Explorative ($\beta = 3$). The proximity-based acquisition function consistently achieves a good balance between the regret and the reliance on high-fidelity usage. In contrast, MF-GPR-UCB exhibits greater inconsistency across the exploration modes. While fidelity-weighted strategy attains the best convergence, it significantly relies on high-fidelity usage.}
      \label{fig:himmelblau_performance}
\end{figure}

\noindent For the Himmelblau function (Figure \ref{fig:himmelblau_performance}), on average, across different exploration strategies, the proximity-based method strikes a good balance between the regret achieved and the percentage of high-fidelity evaluations. MF-GPR-UCB exhibits high variance for the regret achieved for exploitation and adaptive exploration,  indicating inconsistency. The fidelity-weighted method consistently uses a significantly higher percentage of high-fidelity evaluations for all exploration strategies. 

\begin{figure}[H]
     \centering
     \begin{subfigure}[b]{0.32\textwidth}
         \centering
         \includegraphics[width=\textwidth]{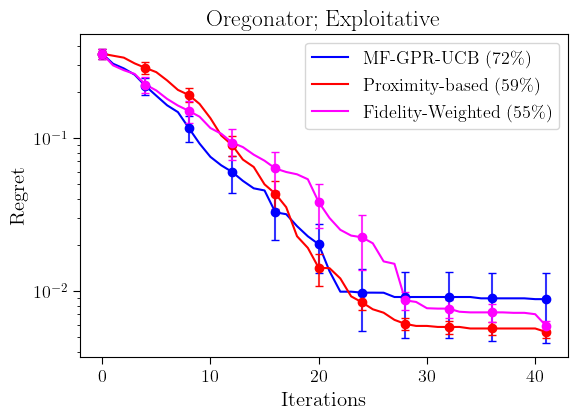}
     \end{subfigure}
     \hfill
     \begin{subfigure}[b]{0.32\textwidth}
         \centering
         \includegraphics[width=\textwidth]{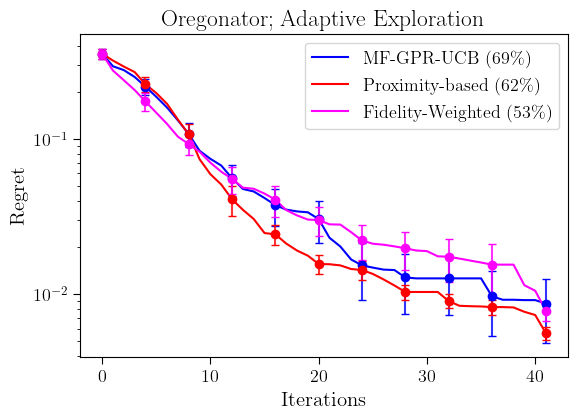}
     \end{subfigure}
     \hfill
     \begin{subfigure}[b]{0.32\textwidth}
         \centering
         \includegraphics[width=\textwidth]{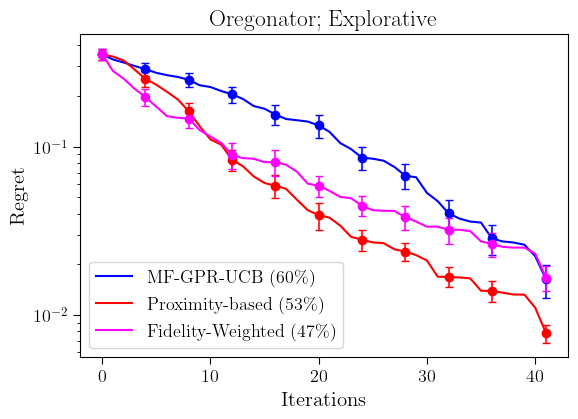}
     \end{subfigure}
     
     \caption{\textbf{Oregonator 2D.} (a) Exploitative ($\beta=0.5$) (b) Adaptive exploration (c) Explorative ($\beta = 3$). The fidelity-weighted strategy exhibits the least reliance on high-fidelity usage, while the proximity-based strategy exhibits better convergence, albeit with a slight increase in high-fidelity usage. In contrast, the MF-GPR-UCB struggles to reduce the regret even with a higher percentage of high-fidelity evaluations.  }
      \label{fig:oreg_performance}
\end{figure}

\noindent For the optimization problem to detect Hopf bifurcation parameters of the Oregonator model, all three acquisition functions have similar regret performance (Figure \ref{fig:oreg_performance}). MF-GPR-UCB consistently uses more high-fidelity evaluations compared to the other two. The fidelity-weighted method uses the lowest percentage of high-fidelity evaluations, while the proximity-based method achieves a superior regret with a slightly higher percentage of high-fidelity evaluations.

\begin{figure}[H]
     \centering
     \begin{subfigure}[b]{0.32\textwidth}
         \centering
         \includegraphics[width=\textwidth]{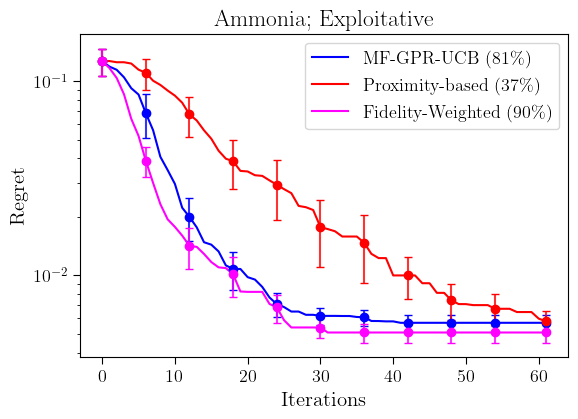}
     \end{subfigure}
     \hfill
     \begin{subfigure}[b]{0.32\textwidth}
         \centering
         \includegraphics[width=\textwidth]{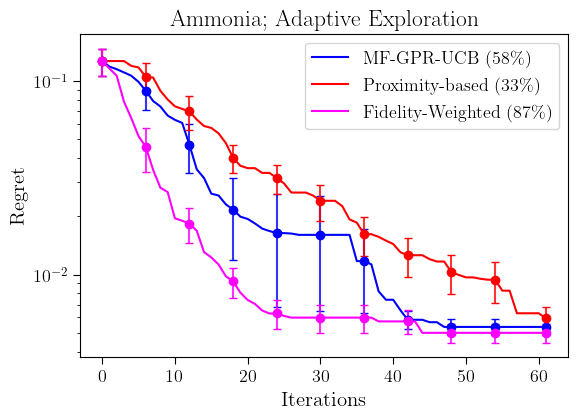}
     \end{subfigure}
     \hfill
     \begin{subfigure}[b]{0.32\textwidth}
         \centering
         \includegraphics[width=\textwidth]{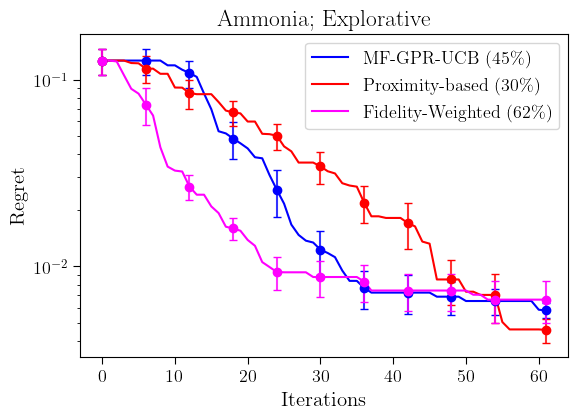}
     \end{subfigure}
     
     \caption{\textbf{Ammonia 2D.} (a) Exploitative ($\beta=0.5$) (b) Adaptive exploration (c) Explorative ($\beta = 3$). The proximity-based acquisition function consistently converges to the optimum using a lower percentage of high-fidelity evaluations. In contrast, MF-GPR-UCB exhibits inconsistent high-fidelity usage, and the fidelity-weighted strategy relies heavily on high-fidelity usage.}
      \label{fig:ammonia_performance}
\end{figure}

\noindent For the dynamic ammonia catalysis model (Figure \ref{fig:ammonia_performance}), we see that the proximity-based acquisition function outperforms the others by achieving similar regret levels using a significantly lower percentage of high-fidelity evaluations. We also note that the low-fidelity optimum is quite inaccurate, and distant from the high-fidelity model one, which necessitates a larger number of high-fidelity evaluations by the fidelity-weighted method.

\begin{figure}[H]
     \centering
     \begin{subfigure}[b]{0.32\textwidth}
         \centering
         \includegraphics[width=\textwidth]{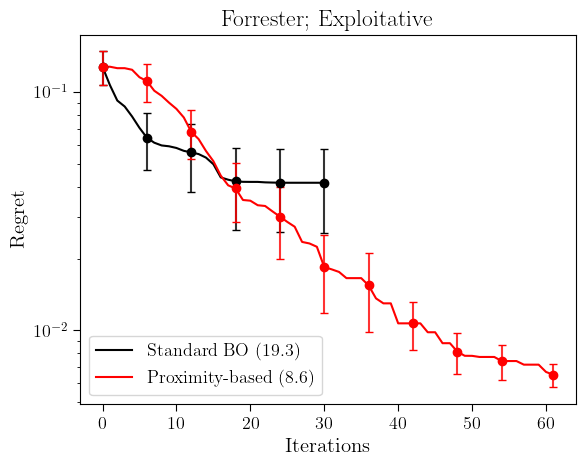}
     \end{subfigure}
     \hfill
     \begin{subfigure}[b]{0.32\textwidth}
         \centering
         \includegraphics[width=\textwidth]{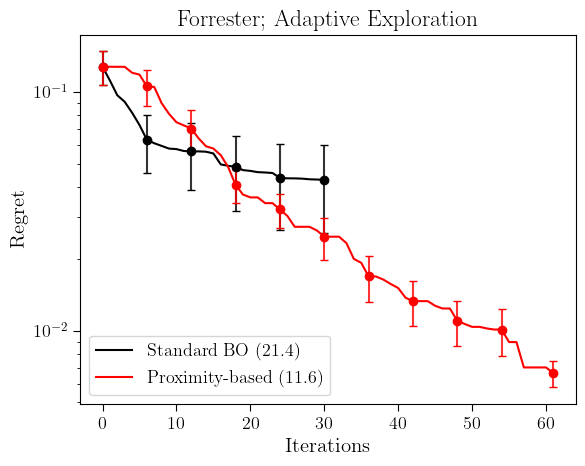}
     \end{subfigure}
     \hfill
     \begin{subfigure}[b]{0.32\textwidth}
         \centering
         \includegraphics[width=\textwidth]{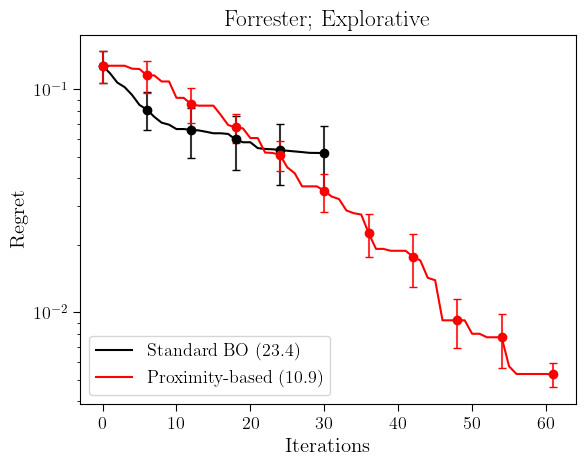}
     \end{subfigure}
     
     \caption{\textbf{Ammonia 2D.} (a) Exploitative ($\beta = 0.5$) (b) Adaptive exploration (c) Explorative ($\beta = 3$). For all three modes, the proximity-based acquisition function consistently converges to the global optimum using a lower number of high-fidelity evaluations compared to standard BO. The value in the parenthesis denotes the average number of high-fidelity evaluations accessed.}
      \label{fig:amm_bo_comparison}
\end{figure}

\noindent Figure \ref{fig:amm_bo_comparison} shows that for the ammonia catalysis model, ``standard" BO is outperformed by the proximity-based multi-fidelity BO. There is high variance in the best solution achieved by standard BO. Since both algorithms were initialized with the same number of high-fidelity evaluations, the standard BO starts with a very sparse dataset, making the performance of the algorithm heavily dependent on initialization. The proximity-based multi-fidelity BO exhibits a tighter bound on the final regret achieved, indicating good performance for all initializations. Additionally, the multi-fidelity technique uses less high-fidelity evaluations, as the search space is explored mostly using low-fidelity evaluations.

\subsection{Cost ratio parameter comparison}
\begin{figure}[H]
     \centering
     \begin{subfigure}[b]{0.32\textwidth}
         \centering
         \includegraphics[width=\textwidth]{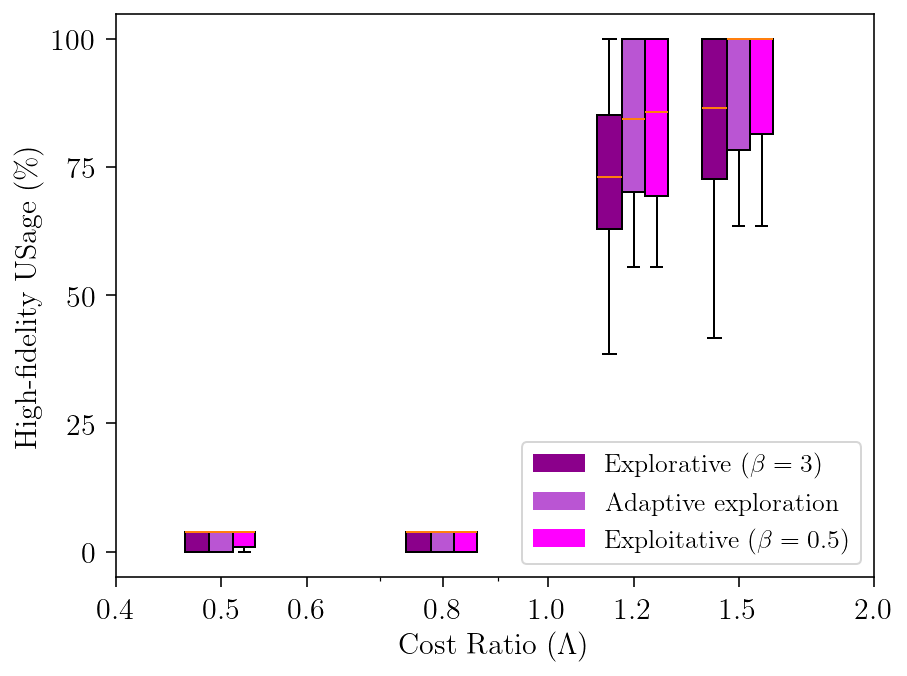}
     \end{subfigure}
     \hfill
     \begin{subfigure}[b]{0.32\textwidth}
         \centering
         \includegraphics[width=\textwidth]{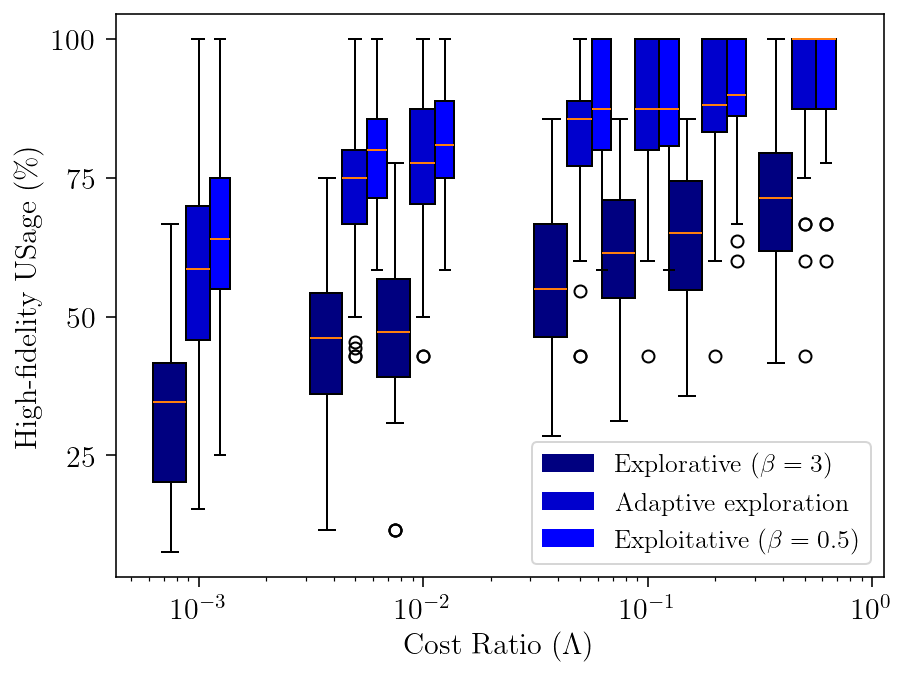}
     \end{subfigure}
     \hfill
     \begin{subfigure}[b]{0.32\textwidth}
         \centering
         \includegraphics[width=\textwidth]{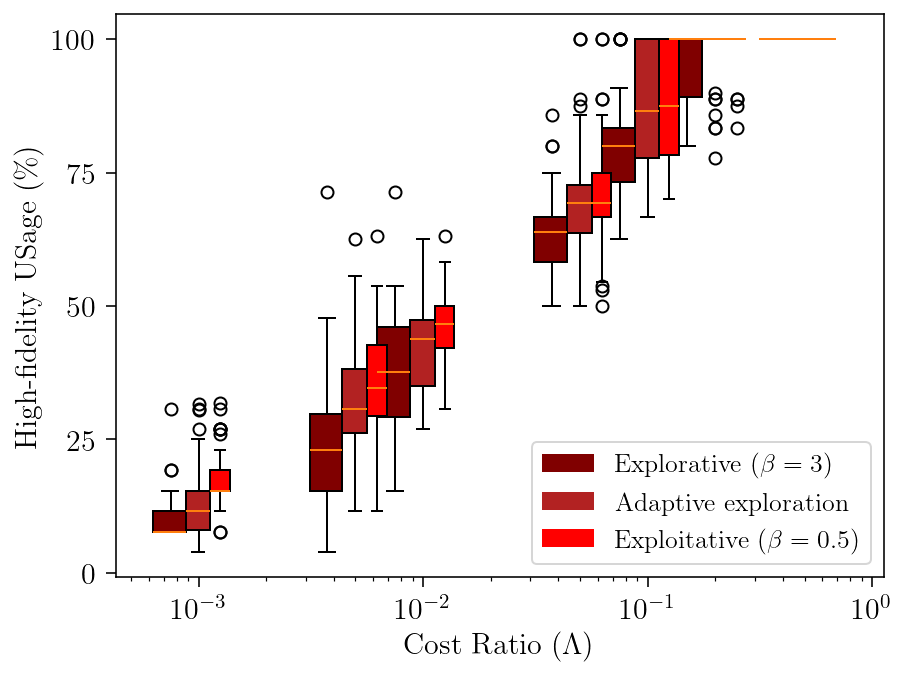}
     \end{subfigure}
     
     \caption{\textbf{Toy-enzyme 1D}. (a) Fidelity-weighted (b) MF-GPR-UCB (c) Proximity-based. The fidelity-weighted strategy exhibits a sharp increase in high-fidelity usage around $\Lambda=1$. In contrast, the other strategies show a smoother and more predictable trends, with the proximity-based strategy exhibiting tighter box plots} 
     \label{fig:cost_enzyme}
     \end{figure}

\begin{figure}[H]
     \centering
     \begin{subfigure}[b]{0.32\textwidth}
         \centering
         \includegraphics[width=\textwidth]{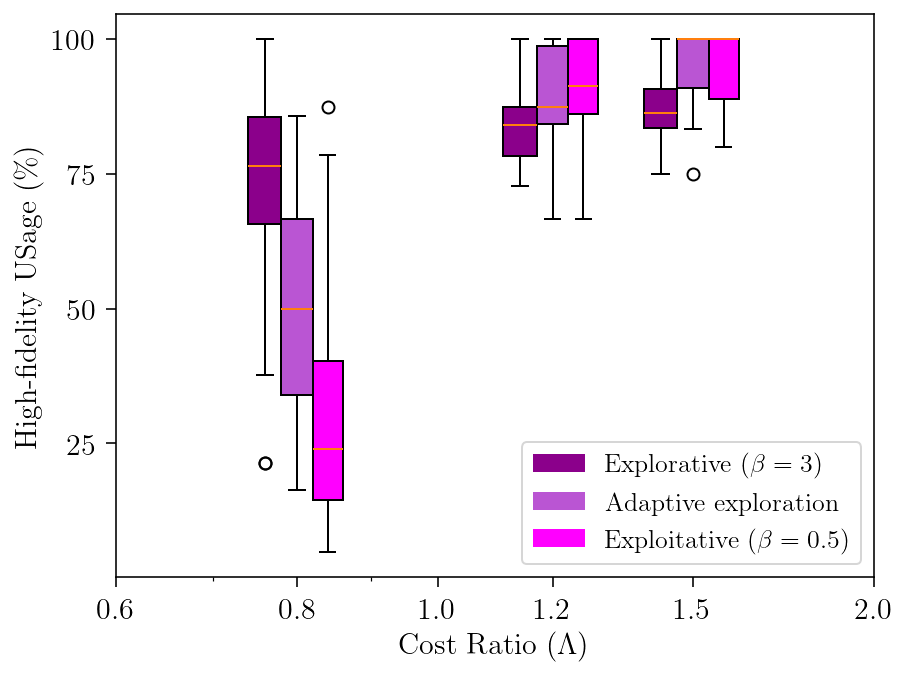}
     \end{subfigure}
     \hfill
     \begin{subfigure}[b]{0.32\textwidth}
         \centering
         \includegraphics[width=\textwidth]{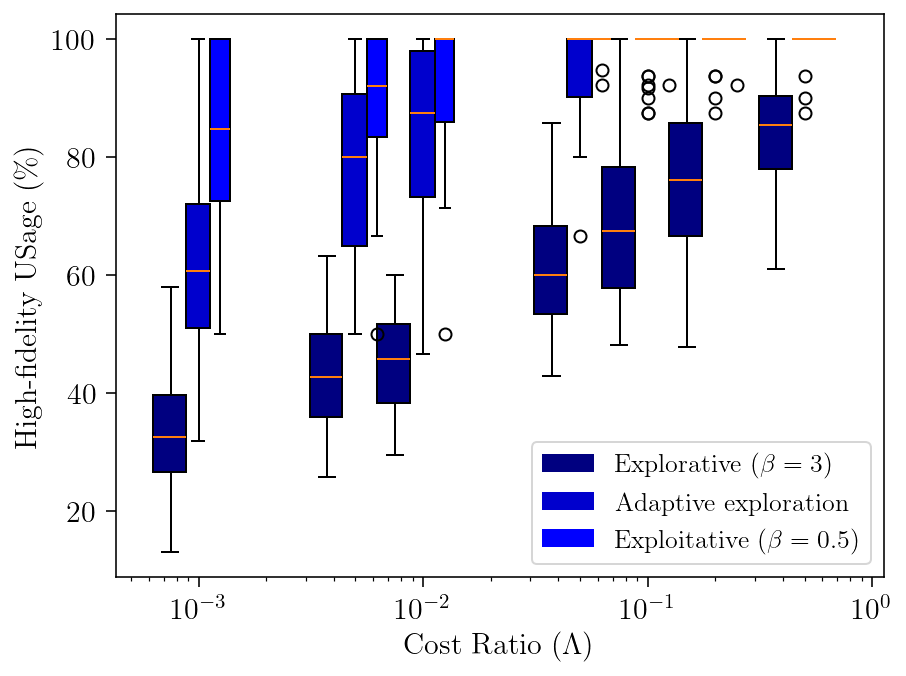}
     \end{subfigure}
     \hfill
     \begin{subfigure}[b]{0.32\textwidth}
         \centering
         \includegraphics[width=\textwidth]{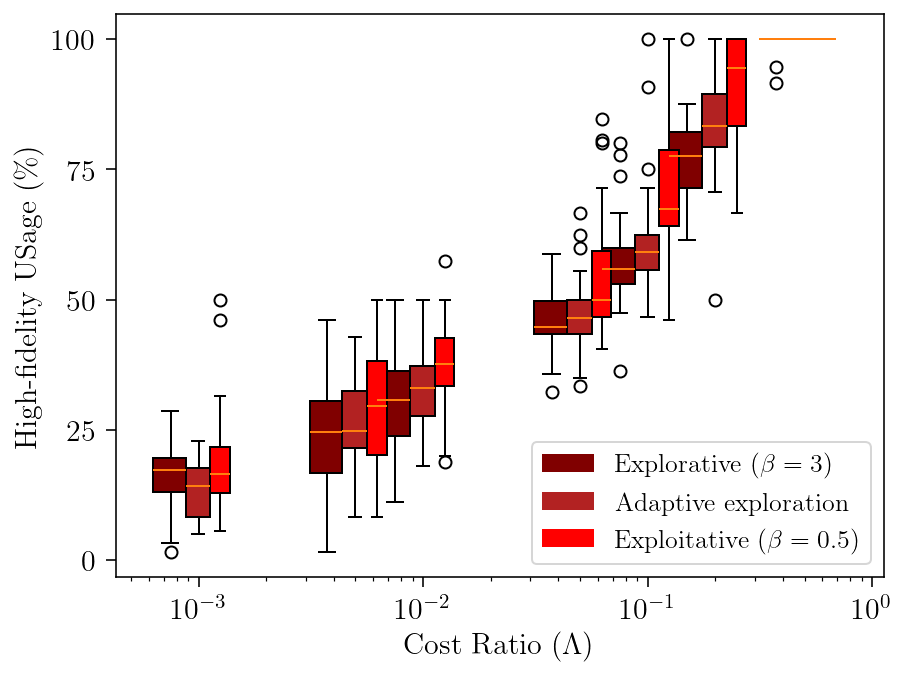}
     \end{subfigure}
     
     \caption{\textbf{Ammonia 2D}. (a) Fidelity-weighted (b) MF-GPR-UCB (c) Proximity-based. The fidelity-weighted strategy exhibits wider box plots exhibiting inconsistency in the high-fidelity usage. The other strategies show a smoother and more predictable trends, with the proximity-based strategy showing tighter distributions. Notably, the exploitative mode for the MF-GPR-UCB strategy exhibit elevated high-fidelity usage even at low $\Lambda$ values.}
     \label{fig:cost_ammonia}
     \end{figure}

\begin{figure}[H]
     \centering
     \begin{subfigure}[b]{0.32\textwidth}
         \centering
         \includegraphics[width=\textwidth]{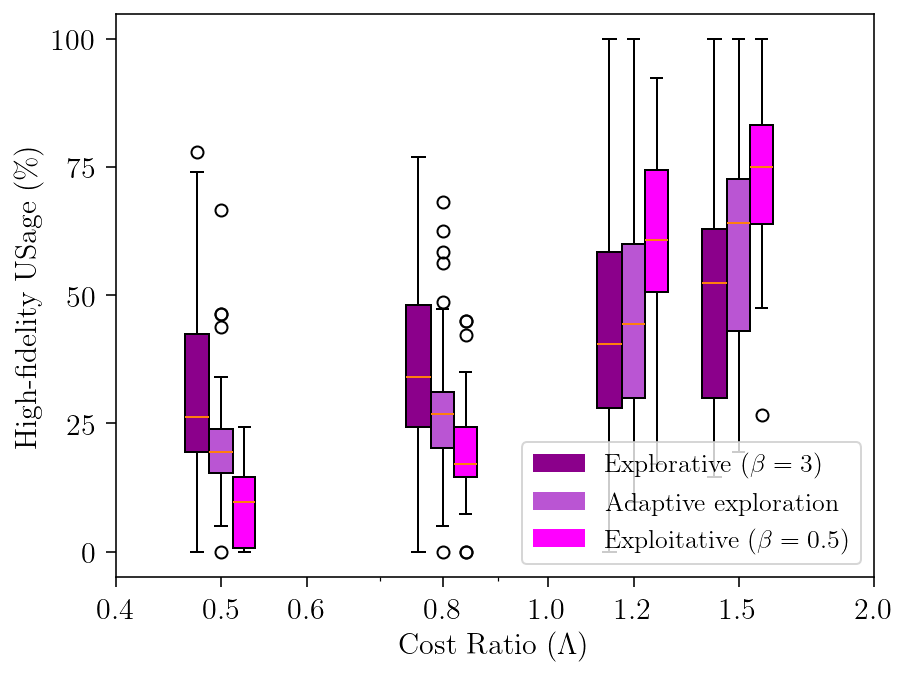}
     \end{subfigure}
     \hfill
     \begin{subfigure}[b]{0.32\textwidth}
         \centering
         \includegraphics[width=\textwidth]{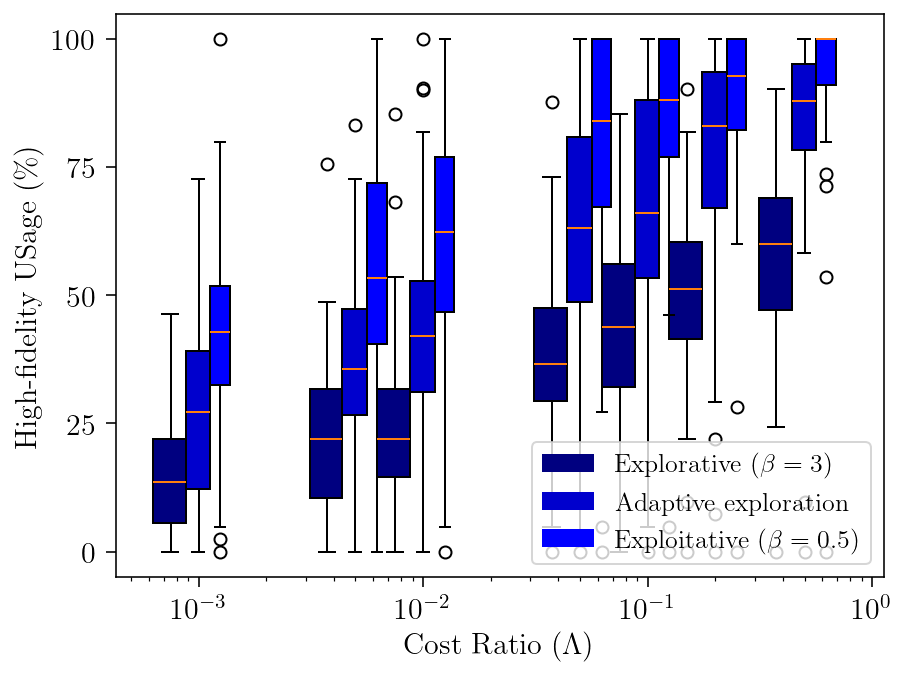}
     \end{subfigure}
     \hfill
     \begin{subfigure}[b]{0.32\textwidth}
         \centering
         \includegraphics[width=\textwidth]{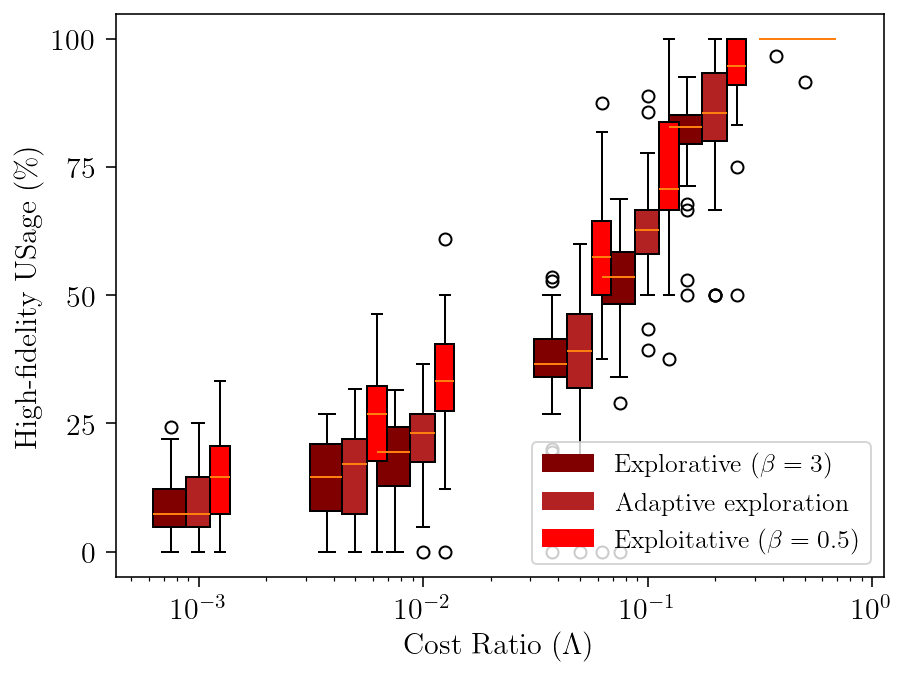}
     \end{subfigure}
     
     \caption{\textbf{Oregonator 2D}. (a) Fidelity-weighted (b) MF-GPR-UCB (c) Proximity-based. In comparison to the previous test functions, the fidelity-weighted strategy displays a smoother trend in high-fidelity usage. Although the MF-GPR-UCB strategy exhibits a smoother trend for high-fidelity usage, the box plots are noticeably wider. In contrast, the proximity-based strategy exhibits a smooth and consistent trend with tighter box plots.}
     \label{fig:cost_oreg}
     \end{figure}

We compare the effect of the cost ratio parameter ($\Lambda = \lambda_1/\lambda_2$) across different acquisition functions. To achieve a more comprehensive understanding, we repeat the experiments for different values of the exploration-exploitation parameter $\beta$ as summarized in Table \ref{tab:forrester_performance}. To quantify the \textit{tunability} of the different optimization strategies, we plot the high-fidelity usage against the respective cost ratio for three different exploration modes (Figures \ref{fig:cost_enzyme}, \ref{fig:cost_ammonia} and \ref{fig:cost_oreg}). All  optimization runs are averaged over up to 50 different LHS initializations for each  cost ratio value. We show results across three different benchmark problems--- Toy enzyme, ammonia model, and the Oregonator. Additional results and extended figures for the rest of the test functions can be found in \ref{app:cost}. The box plots represent the high-fidelity usage for various cost-ratio parameters, with different shades within each figure representing distinct modes of exploitation ($\beta=0.5$), exploration ($\beta=3$) and adaptive exploration.

\noindent We observe that the fidelity-weighted method exhibits sharp sensitivity to the cost ratio. This is clearly visible for the toy enzyme model and the ammonia model, where the high-fidelity usage increases abruptly. Implementing this for practical problems not previously encountered would require more fine-tuning than other acquisition functions. On the other hand, the cost-ratio for the other two acquisition functions is confined between 0 and 1, providing easier control of the fraction of high-fidelity evaluations; this is key for computationally expensive problems, where we would like to minimize high-fidelity evaluations. The MF-GPR-UCB exhibits more consistent control and predictability when compared with the fidelity-weighted strategy. As the cost-ratio increases, there rise in high-fidelity usage is smooth and consistent, suggesting easier tuning of the parameter to control high-fidelity usage. However, the variance bounds remain quite high, especially in low cost-ratio regimes. We also notice that for certain problems, the high-fidelity usage is still high for near zero cost ratio values. 

The proximity-based acquisition function exhibits better overall hyperparameter controllability and stands out as the most predictable and tunable across all problems. In addition to the smooth and consistent rise in high-fidelity usage, the box plots are more tightly grouped, indicating robustness across test functions and exploration modes. Additionally, apart from the fidelity-weighted method, we observe a consistent trend across the exploration modes for all the test functions: The exploitative strategy accesses more high-fidelity evaluations, while the explorative strategy tends to use the least. This is intuitive, as the exploitative strategy tends to select points closer to previously evaluated low-fidelity samples, increasing the likelihood of triggering high-fidelity evaluations.

\section{Conclusions}
We have explored the application of multi-fidelity models in Bayesian Optimization. The fidelity-weighted method generally performs well across most of the benchmark functions, though it tends to rely heavily on high-fidelity evaluations. MF-GPR-UCB strikes a good balance between high-fidelity usage and the regret achieved in most cases, although it shows inconsistency in certain cases  like the exploitative strategy for the Himmelblau function. The proximity-based acquisition function consistently demonstrates good performance on all the cases, with a good balance of high-fidelity usage. Notably, we observe good performance for the Forrester function---where the low-fidelity optimum ``tricks" the multi-fidelity model into converging to the local optimum nearby. The same acquisition function also proves to be highly effective in optimizing the ammonia model with a significantly lower percentage of high-fidelity evaluations. Additionally, the parametric analysis showed that tuning the cost-parameter of the fidelity-weighted acquisition function is heavily problem-dependent. Nevertheless, the simpler and more intuitive acquisition functions provide better control of the high-fidelity usage for optimization. We found that the proximity-based acquisition function exhibits better control among all the acquisition functions tested.

Analysis of the evaluation strategies reveals that since the Fidelity-Weighted method uses two separate acquisition functions, there is no mutual information exchange between the two -- only the high-fidelity GP ``learns" from the low-fidelity data. The low-fidelity GP remains the same, and this causes the low-fidelity acquisition function to end up pointing to the low-fidelity optimum at all times. This means that the algorithm might get stuck at the low-fidelity optimum if this is located far from the high-fidelity optimum, as was observed in the Forrester function. The design of the other acquisition functions prohibits the algorithm from getting stuck near low-fidelity optima, as the conditions in the algorithms necessitate a switch to high-fidelity if/when low-fidelity is sampled ``too often". Kandasamy et al. proposed the idea of a combined acquisition function for better information exchange when the high-fidelity surrogate is a standard GP. However, in the case of multi-fidelity GP, this information exchange is inherent for the high-fidelity surrogate. Moreover, this strategy assumes prior knowledge of the error bound between the fidelity levels to guide fidelity selection.
Hence, we can simply converge to selecting the proximity-based acquisition strategy which optimizes a single high-fidelity acquisition function, where fidelity selection is guided by a single cost-ratio parameter.

For future work, we plan to incorporate Global Optimization computations for our multi-fidelity acquisition functions.  Commonly used local optimizers may inadequately optimize these acquisition functions, which often have multimodal landscapes, thus reducing the overall efficiency of BO. Recent advances in GP optimization~\cite{schweidtmann2021} make global optimization of GPs now feasible, and while related works theorize about potential benefits~\cite{wilson2018maximizing}, these have yet to be tested extensively in practice. However, Georgiou et al. \cite{georgiou2025deterministic} leveraged the open-source deterministic solver MAiNGO (McCormick-based Algorithm for mixed-integer Nonlinear Global Optimization) \cite{bongartz2018maingo} which employs a reduced-space formulation to conclude global optimization exhibits faster convergence for an exploitative setting.
By utilizing this optimization approach in the context of our multi-fidelity framework, we aim to further explore the use of global optimization for GPs.

\appendix

\section{Nomenclature}
\label{sec:Nomenclature} 

\begin{table}[H]
\caption{Summary of the main symbols, notations, and abbreviations used in this work.}
\centering
\begin{tabular}{c l}
\toprule
 Notation & Description  \\
 \toprule
$\mathcal{GP}$ or GP & Gaussian Process\\
$\mathcal{MFGP}$ & Multi-fidelity Gaussian Process\\
$s$ & fidelity level\\
$\kappa$ & Covariance kernel\\
$\bm{K}$ & Covariance Matrix\\
$\mu$ & $\mathcal{GP}$ or $\mathcal{MFGP}$ mean prediction\\
$\sigma^2$ & $\mathcal{GP}$ or $\mathcal{MFGP}$ variance prediction\\
$\delta$ & Correction GP\\
$\rho$ & $\mathcal{MFGP}$ scaling factor\\
$\mathcal{D}$ & Data\\
$\alpha$ & Acquisition function\\
$\beta$ & Exploration parameter\\
$\zeta$ & Multi-fidelity UCB error bound\\
$\gamma$ & Multi-fidelity UCB threshold\\
$\lambda$ & Cost weight\\
$\Lambda$ & Cost ratio\\
MMD & Average sum of minimum distances\\
MF-GPR & Multi-fidelity Gaussian Process Regression\\
BO & Bayesian Optimization\\
UCB & Upper Confidence Bound\\
TOF & Turnover Frequency\\
\bottomrule
\end{tabular}
\label{table:nomenclature}
\end{table}

\section{Gaussian Process Regression \label{app:gp}}

\subsection{Gaussian Processes}
\noindent Gaussian processes \cite{GP} are a collection of random variables, any finite number of which have a joint Gaussian distribution. Gaussian processes are represented by a mean function ($m$) and a covariance function ($\kappa$). Parameterized by a vector of hyperparameters $\theta$, the covariance function specifies the covariance between two locations $\bm{x}$ and $\bm{x}'$. An example of a covariance function is the squared exponential function (\ref{sqexp}). Assuming that no prior information on the underlying objective function is available, GPs are typically initialized as zero-mean GPs ($f(\bm{x}) \sim \mathcal{GP}(0, \kappa_{low}(\bm{x}, \bm{x'})$). The use of such GP-based models is enabled by python packages such as BoTorch \cite{balandat2020botorch}, Trieste \cite{picheny2023trieste}, Emukit \cite{emukit2019, emukit2023} and GPyOpt \cite{gpyopt2016}. 
\begin{equation}
    f \sim \mathcal{GP}(m(\bm{x}), \kappa(\bm{x}, \bm{x'},\theta))
\end{equation}

   \begin{equation}\label{sqexp}
       \kappa(\bm{x},\bm{x}') = \sigma_f^2 exp \left({-\frac{\norm{\bm{x}-\bm{x}'}}{2l^2}} \right) 
       \quad \bm{\theta}=\{\sigma_f, l \}
   \end{equation}

\begin{figure}[H]
     \centering
     \begin{subfigure}[b]{0.49\textwidth}
         \centering
         \includegraphics[width=\textwidth]{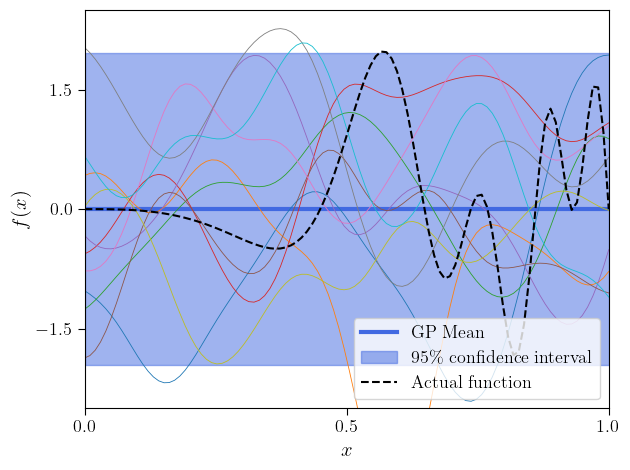}
     \end{subfigure}
     \hfill
     \begin{subfigure}[b]{0.49\textwidth}
         \centering
         \includegraphics[width=\textwidth]{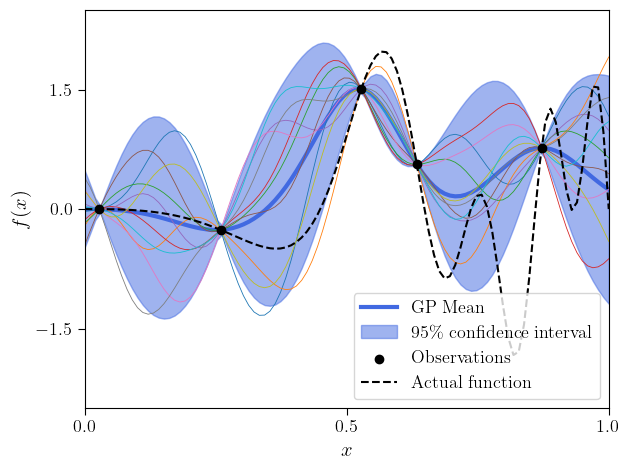}
     \end{subfigure}
     \hfill
     \caption{\textbf{Gaussian Processes.} (a) The Gaussian process is initialized using a zero mean prior, representing uncertainty in the absence of data points. (b) The posterior prediction of the GP trained with observed data points. The sample trajectories illustrate realizations from the prior or posterior distributions.}
\end{figure}

\noindent The hyperparameters of a GP are generally optimized by maximizing the marginal likelihood using optimization techniques like BFGS with multiple restarts.

\begin{equation}\label{loglik}
       log P(\mathcal{D}|\bm{\theta}) = -\frac{1}{2}log|\bm{K}| - \frac{1}{2}\bm{y}^T\bm{K}^{-1}\bm{y} - \frac{N}{2}log(2\pi)
   \end{equation}

\subsection{Gaussian Process Regression}

\noindent With observed data $\mathcal{D} = \{x_i, y_i\}_{i=1}^{N}$, Gaussian Process regression computes the posterior probability $p(f(\bm{x})|\mathcal{D})$ at each point $\bm{x}$ as a Gaussian with a mean and covariance,
\begin{equation}\label{gpmean}
    \mu(\bm{x}) = \kappa(\bm{x}, \bm{X})\bm{K}^{-1}\bm{y}
\end{equation}

\begin{equation}\label{gpvar}
    \sigma^2(\bm{x}) = \kappa(\bm{x},\bm{x}) - \kappa(\bm{x},\bm{X})\bm{K}^{-1}\kappa(\bm{X},\bm{x})
\end{equation}

\noindent where $\bm{X}$ is the set of design locations and $\bm{y}$ is the corresponding set of expensive function evaluations. 

\section{Acquisition Functions \label{app:acq}}
\noindent An Acquisition Function is a (hopefully cheap to optimize) function which balances the exploitation of known good state space regions based the GP mean predictions and the exploration of uncertain regions with higher GP variance. An appropriate extremum of this tractable acquisition function, $\alpha(x)$ gives us the next best candidate. 

\subsection{Upper Confidence Bound}
\begin{equation}\label{UCB}
    \alpha = \mu(\bm{x}) + \beta^{1/2} \sigma (\bm{x})
\end{equation}
UCB \cite{UCb} has a direct tradeoff between expected performance captured by $\mu(x)$ and the uncertainty captured by $\sigma(x)$. $\beta$ is a hyperparameter which controls this tradeoff. Lower values of $\beta$ lead to an {\em exploitative} approach, and higher values lead to an {\em explorative} approach

\subsection{Expected Improvement}
\noindent The expected improvement (EI) acquisition function quantifies the expected value of improvement over the current best solution, denoted as $\max(f^* - f(\bm{x}), 0)$, where $f^*$ is the best solution found so far.

\begin{equation}
    EI(\bm{x}) = \mathbb{E}[\max(f^* - f(\bm{x}), 0)]
\end{equation}

\noindent Since $f(\bm{x}) \sim \mathcal{N}(\mu(\bm{x}), \sigma(\bm{x})^2)$, the EI has a closed-form expression,

\begin{equation}
    \text{EI}(\bm{x}) = (f^* - \mu(\bm{x})) \cdot \Phi(Z) + \sigma(\bm{x}) \cdot \phi(Z)
\quad \text{with} \quad Z = \frac{f^* - \mu(\bm{x})}{\sigma(\bm{x})}
\end{equation}

\noindent Where $\phi(Z)$ and $\Phi(Z)$ are the probability density function (PDF) and cumulative density function (CDF) of the standard normal distribution evaluated at Z. The modified weighted EI is obtained using an exploration hyperparameter $\beta$,

\begin{equation}
    \text{$EI_w$}(\bm{x}) = (f^* - \mu(\bm{x})) \cdot \Phi(Z) + \beta \sigma(\bm{x}) \cdot \phi(Z)
\end{equation}

\section{Implementation details for Multi-fidelity GP Regression}
\label{app:mfgpr}
\noindent 
\textit{Initialization}: The low-fidelity data for all test cases was initialized using Latin hypercube sampling. The initialization of multi-fidelity GPs demands that the samples of the high-fidelity data are a subset within the low-fidelity data.\\ 
\textit{Kernel}: We used the squared-exponential kernel in all of our test cases. The kernel is initialized and updated at each iteration through kernel hyperparameter optimization.\\
\textit{Choice of $\beta$}: In addition to exploitative ($\beta$ = 0.5 and $\beta$ = 1) and explorative choices ($\beta$ = 3 and $\beta$ = 5), we adopted the adaptive exploration proposed by Kandasamy et al. \cite{mfucb} that captures dependencies on the dimensionality of the problem ($d$) and the number of iterations ($t$).

\begin{equation}\label{varyingbeta}
    \beta^2 = 0.2d\log(2t)
\end{equation}

\begin{figure}[H]
      \centering
      \includegraphics[width=0.45\linewidth]{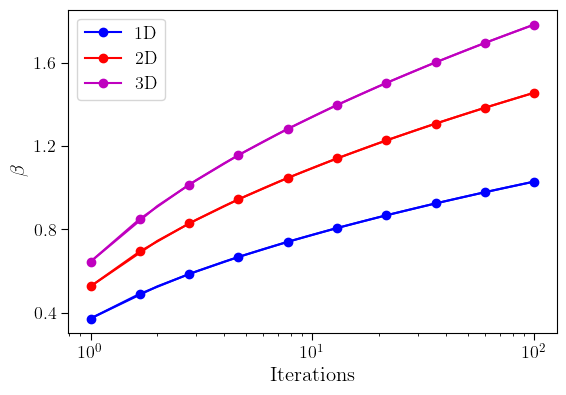}
      \caption{\textbf{Adapative exploration.} The variation of the exploration parameter ($\beta$) plotted against the number of iterations for different problem dimensions.}
      \label{fig:enzyme}
\end{figure}

\noindent \textit{Acquisition Optimizer}: The acquisition function(s) are optimized using L-BFGS with multiple-restarts. For the fidelity-weighted strategy, which involves separate acquisition functions for each fidelity level, both of them are optimized independently, and the best location is selected by comparing these optima.\\
\textit{Bayesian Optimization}: As the optimization progresses, the algorithm may choose to perform high-fidelity evaluations at locations not previously sampled with low-fidelity. To satisfy the nested design requirement, the multi-fidelity GP is trained by assuming that the low-fidelity value at the new point is approximated as $f_{low} = \rho\mu_{low}(x_{new})$, where $\mu_{low}$ is the mean prediction of the low-fidelity GP and $\rho$ is the scaling factor. The hyperparameters were updated after every iteration. Alternatively, one could also choose to update them less frequently to reduce the computational overhead.

\section{Test Functions}\label{app:tf}
\subsection{Forrester function}
\noindent We minimize this one-dimensional synthetic test function \cite{forrester}, where the low-fidelity function is a slightly modified version of the high-fidelity function. Notably, the low-fidelity optimum is significantly distant from the high-fidelity optimum. The high- and low-fidelity functions are,

\begin{equation}\label{eqn:forrester}
    \begin{aligned}
    f_{high}(x) &= (6x - 2)^2 \sin (12x-4),\\
    f_{low}(x) &= \frac{1}{2}f_{high}(x) + 10\left(x - \frac{1}{2}\right) - 5.
    \end{aligned}   
\end{equation}

\begin{figure}[H]
      \centering
      \includegraphics[width=0.45\linewidth]{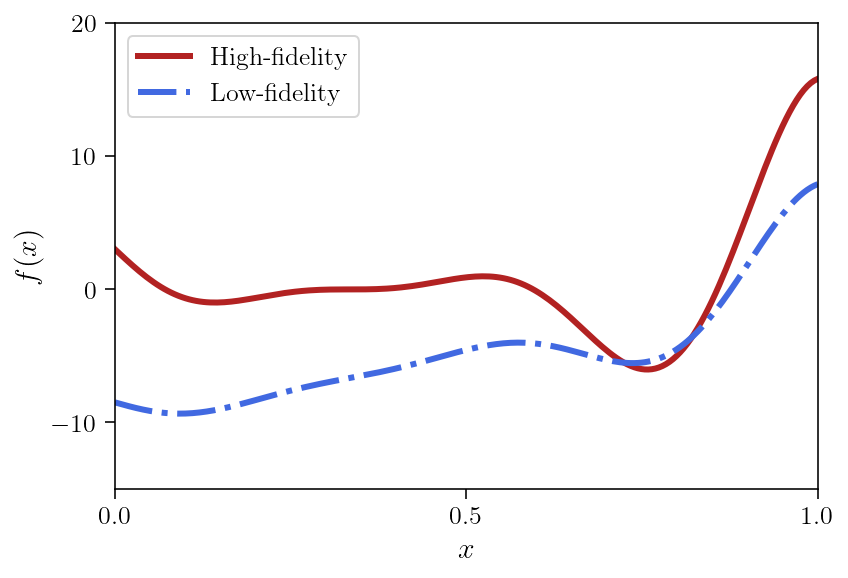}
      \caption{\textbf{Multi-fidelity Forrester Function.} Low-fidelity and high-fidelity functions.}
      \label{fig:forrester}
  \end{figure}

\subsection{Bohachevsky Function}
\noindent We minimize two-dimensional synthetic test function \cite{bohachevskyhimmelblau} in which the low-fidelity function is a slight modification of the high-fidelity function. The domain is the two-dimensional cube $\mathcal{X} = [-5, 5]^2$. The high- and low-fidelity functions are,

\begin{equation}\label{eqn:bohachevsky}
    \begin{aligned}
    f_{high}(x_1, x_2) &= x_1^2 + 2x_2^2 - 0.3\cos (3\pi x_1) - 0.4\cos (4\pi x_2) + 0.7,\\
    f_{low}(x_1, x_2) &= f_{high}(0.7x_1, x_2) + x_1x_2 - 12.
    \end{aligned}   
\end{equation}

\subsection{Himmelblau Function}
\noindent We minimize this two-dimensional synthetic test function \cite{bohachevskyhimmelblau} in which the low-fidelity function is a slight modification of the high-fidelity function. This test function has 4 global optima which are found by solving this as a maximizing problem. The domain is the two-dimensional cube $\mathcal{X} = [-4, 4]^2$. The high- and low-fidelity functions are,

\begin{equation}\label{eqn:himmelblau}
    \begin{aligned}
    f_{high}(x_1, x_2) &= (x_1^2 + x_2 - 11)^2 + (x_2^2 + x_1 - 7)^2,\\
    f_{low}(x_1, x_2) &= f_{high}(0.5x_1, 0.8x_2) + x_2^3 - (x_1+ 1)^2.
    \end{aligned}   
\end{equation}

\subsection{Temperature dependent Oregonator Model}
\noindent The Belousov-Zhabotinsky (BZ) reaction is a classic example of nonlinear chemical reactions that exhibit rich dynamical behavior including periodic temporal oscillations and chaos. BZ reaction involves the oxidation of organic substrates, typically malonic acid, by bromate in an acidic medium catalyzed by metal ions, such as Cerium. Field, Koros and Noyes pioneered the well-known  FKN reaction mechanism \cite{BZ-FKN, BZ-FKN2} by conducting experimental and theoretical studies on the BZ reaction dynamics. Oregonator is a simplified version that can reproduce essential characteristics of the BZ-FKN raection mechanism. The oregonator model consists of three concentration variables involved in five irreversible reactions and a stoichiometric factor ($f$), which denotes the ratio of bromide ions produced to the Cerium ions consumed during a key oxidation step. The system is simplified and rendered dimensionless to obtain the following set of coupled ODEs.
\begin{equation}\label{eqn:oreg_app}
\begin{aligned}
    \varepsilon(T)\dot x &= q(T)ay - xy + ax -x^2\\
    \omega(T)\dot y &= -q(T)ay - xy +fbz\\
    \dot z &= ax - bz
\end{aligned}
\end{equation}

\noindent Here, the model captures the evolution of three species where $x, y$ and $z$ denote $\text{HBrO}_2$, $\text{Br}^-$ and $\text{Ce}^{4+}$ respectively. $a$ and $b$ denote the initial concentrations of malonic acid and sodium bromate. Building on the original oregonator model, Pullela et al. \cite{oregonator(T)} developed a temperature dependent model where the dimensionless parameters $q$, $\varepsilon$, and $\omega$ denote combinations of temperature ($T$) dependent rate constants of the five irreversible reaction steps. After applying QSSA, we develop the multi-fidelity model by varying the parameters $T$ and $f$. The parameter ranges for this multi-fidelity function are $T \in [350, 500K]$ and $f \in [0.5, 2.5]$.

\subsection{Dynamic Ammonia Catalysis Model}

\noindent Wittreich et al. \cite{ammonia} developed an ammonia microkinetic model to computationally evaluate the ammonia synthesis reaction on a Ruthenium catalyst subject to substantial strain oscillations. The kinetic model consists of 16 ODEs that describe 19 elementary steps, including adsorption, desorption, and surface reactions. The binding energies of surface intermediates and transition states were evaluated at different levels of compressive and tensile strain on both terrace and step sites using Density Functional Theory (DFT) to develop the model while also accounting for the effect of surface coverage and surface diffusion. By oscillating the strain using symmetric square waves, Wittreich et al. demonstrated higher catalytic performance\cite{ammonia, programmablecatalyst} under low pressure conditions in a Continuous Stirred-Tank Reactor (CSTR). The ammonia microkinetic model was simulated using a stoichiometric feed (1:3) of $N_2$ and $H_2$, at a fixed temperature of 320 \textdegree C and a reduced pressure of 50 atm (compared to an industrial pressure of 200 atm) on a Ru catalyst with 2\% step sites. The turnover frequency (TOF) is an excellent indicator of catalyst performance and is typically defined as the number of reaction products generated per active catalyst site per unit time. Building on this work, our aim is to discover optimal square waves parameterized by an oscillation frequency ($\nu$) and a duty cycle ($\phi$). The objective is to maximize the period-averaged TOF that can be evaluated at the periodic steady-state (or limit cycle) response to these square waves. The instantaneous TOF can be evaluated using

\begin{equation}\label{eqn:tof_app}
    TOF = \frac{C_{N_2, in} \cdot \dot{Q}_{in} \cdot X_{N_2}}{\rho_{site} \cdot l}.
\end{equation}
Where $\dot{Q}_{in}$ ($cm^3/s$) is the reactor inlet flowrate, $C$ represents concentration and $X_{N_2}$ is the instantaneous conversion of $\text{N}_2$ in the reactor. The parameter $\rho_{site}$ denotes the surface density of the active catalyst sites ($mol/cm^2$), and $l$ represents the catalyst loading per unit reactor volume ($cm^2/cm^3$). We locate the high-fidelity periodic steady-state using Newton-GMRES (\ref{app:nkgmres}), while the low-fidelity steady-state is obtained using a low-accuracy brute-force time integration for 100 time-periods. The newton guess for the high-fidelity solver is generated by performing high-accuracy brute-force time integration for 100 time-periods. The parameter ranges for this multi-fidelity function are $\nu \in [10^2, 10^5 Hz]$ and $\phi \in [0.5, 0.99]$.

\subsubsection{Newton-GMRES for periodic steady-states}
\label{app:nkgmres}

\noindent Given a system of ODEs,
\begin{equation}
    \bm{\dot x} = f(\bm{x}, t; \bm{p})
\end{equation}

\noindent The periodic steady-states of these systems are found by solving a boundary value problem.

\begin{equation}\label{strobores}
    \bm{R}(\bm{x}) = \bm{x} - \bm{S}(\bm{x})
\end{equation}
\noindent Where, 
\begin{equation}\label{strobo}
    \quad \bm{S}(\bm{x}_n) = \int_0^Tf(\bm{x},t;\bm{p})dt \quad  \bm{x}(0) = \bm{x}_n.
\end{equation}

\noindent 
We define the residual $R$ using the stroboscopic map $S$, which represents the state of the system after a full period $T$. The periodic steady-states or limit cycles of such forced dynamical systems are found by locating the roots of the residual equation (\ref{strobores}). To locate these steady-states, we employ Newton's method.

\begin{equation} \label{eqn:newt}
    \bm{J}_R(\bm{x}_n) \cdot \bm{\delta} = -\bm{R}(\bm{x}_n)
\end{equation}

\noindent For higher-dimensional systems, the computational expense of evaluating the full Jacobian can be circumvented using matrix-free methods such as Newton-GMRES \cite{kelley1995iterative, kelley2004newton} which relies on Jacobian-vector products. GMRES is a Krylov subspace method which approximates the solution of a linear system $\bm{A}\bm{\delta} = \bm{b} $ by building Krylov subspaces using matrix-vector products. The solution is approximated as a sum of the form,
\begin{equation}\label{eqn:krylov}
    \bm{\delta}_k = \bm{\delta}_0 + \sum_{k}\gamma_{k}\bm{A}^k\bm{r}_0 \quad\quad \bm{r}_0 = \bm{b} - \bm{A}\bm{\delta}_0.
\end{equation}

\noindent The $k^{th}$ GMRES iterate $\bm{\delta}_k$ is found by minimizing the linear least squares problem $\norm{\bm{b} - \bm{A}\bm{\delta}_k}$ from an initial guess $\bm{\delta}_0$. For each Newton iteration in (\ref{eqn:newt}), a new linear system arises where the  Jacobian-vector products are estimated using directional derivatives.
    \begin{equation} \label{eqn:matvec}
        \bm{J}_R (\bm{x}_n) \cdot \bm{b} = 
        \frac{\bm{R}(\bm{x}_n + \epsilon \frac{\bm{b}}{||\bm{b}||_2}) - \bm{R}(\bm{x}_n)}{\epsilon}
    \end{equation}

\section{Cost Parameter Tunability}\label{app:cost}
\noindent The cost parameter studies for the remaining test functions. We observe similar trends for these test functions as well. The fidelity-weighted strategy is extremely sensitive for the Forrester function. Although there is a consistent rise in the high-fidelity usage for Himmelblau function, the cost ratio assumes negative values and is sensitive around a value of 1. Similar high-fidelity usage trend is observed for the Bohachevsky function. The MF-GPR-UCB strategy tends to have high values of high-fidelity usage across most of the parameter range tested for these three functions. In addition, we observe that the box-plots are wider relative to the other strategies, indicating larger variations. The proximity-based strategy displays less sensitivity and tighter box-plots, demonstrating consistent behavior across all the test functions including the three discussed here.

\begin{figure}[H]
     \centering
     \begin{subfigure}[b]{0.32\textwidth}
         \centering
         \includegraphics[width=\textwidth]{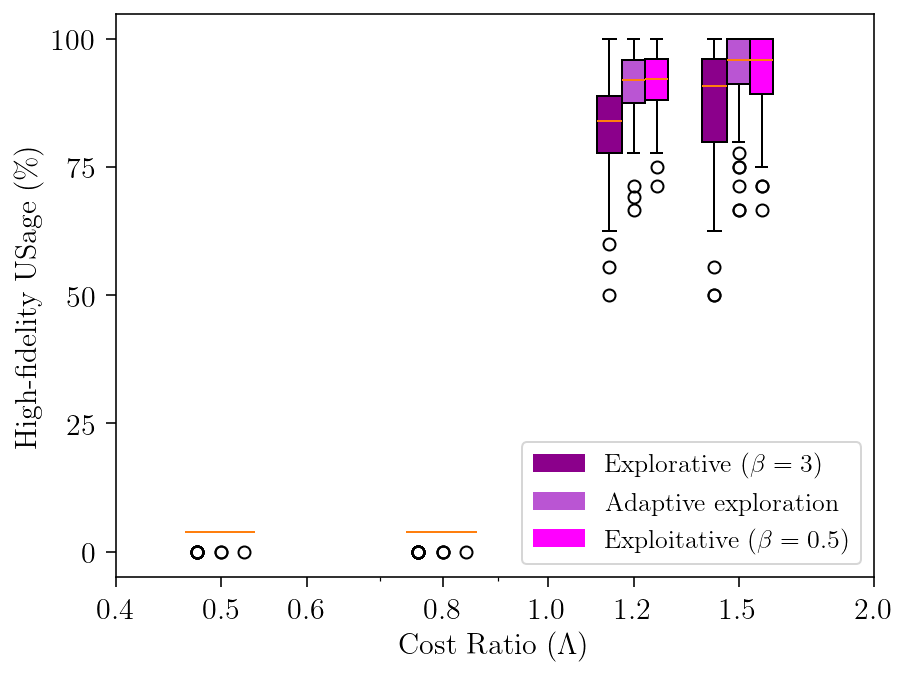}
     \end{subfigure}
     \hfill
     \begin{subfigure}[b]{0.32\textwidth}
         \centering
         \includegraphics[width=\textwidth]{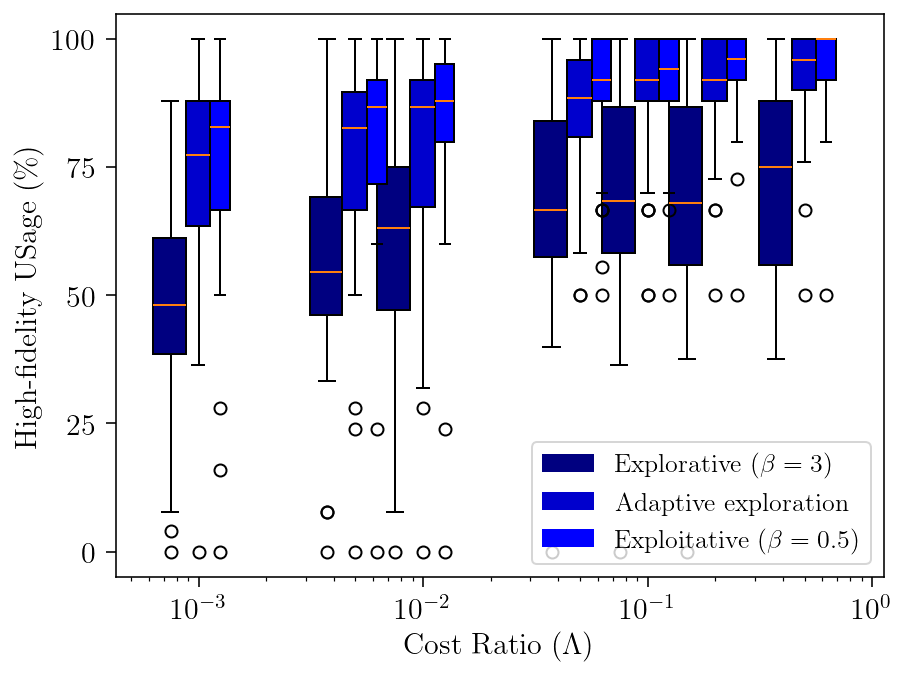}
     \end{subfigure}
     \hfill
     \begin{subfigure}[b]{0.32\textwidth}
         \centering
         \includegraphics[width=\textwidth]{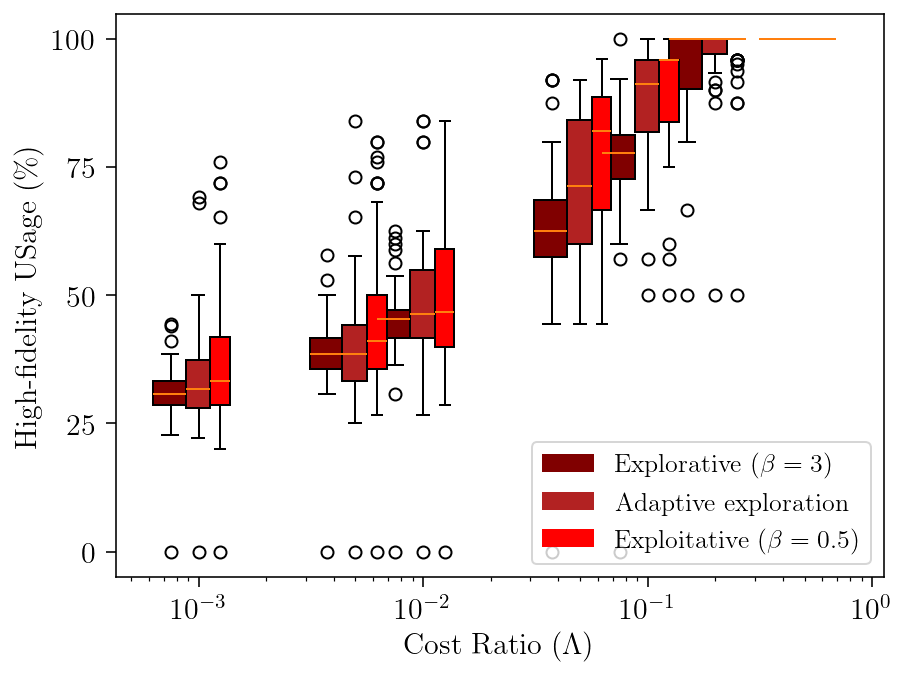}
     \end{subfigure}
     
     \caption{\textbf{Forrester 1D}. (a) Fidelity-weighted (b) MF-GPR-UCB (c) Proximity-based. }
     \label{fig:cost_forrester}
     \end{figure}

\begin{figure}[H]
     \centering
     \begin{subfigure}[b]{0.32\textwidth}
         \centering
         \includegraphics[width=\textwidth]{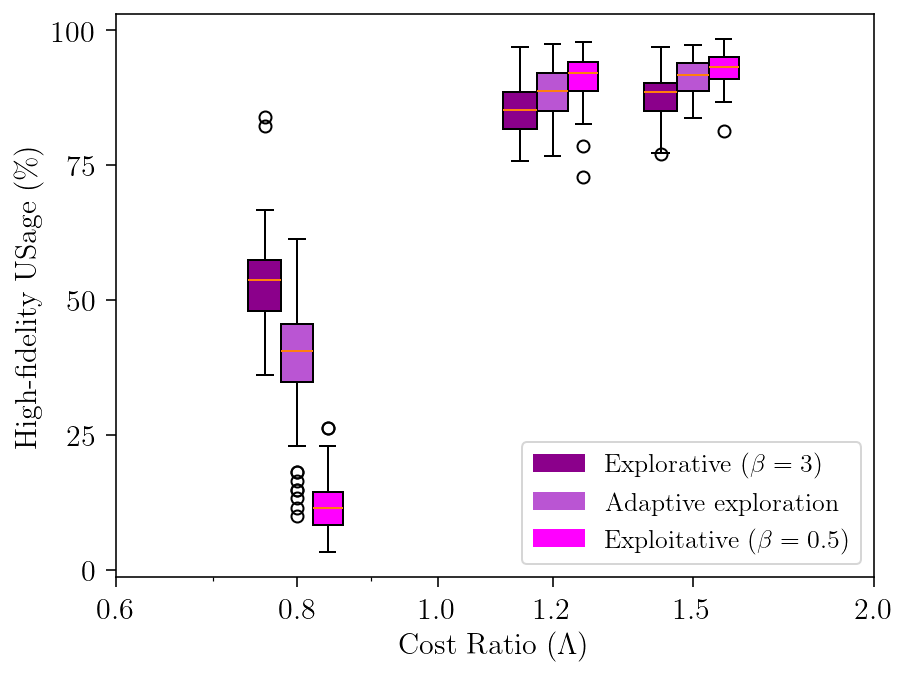}
     \end{subfigure}
     \hfill
     \begin{subfigure}[b]{0.32\textwidth}
         \centering
         \includegraphics[width=\textwidth]{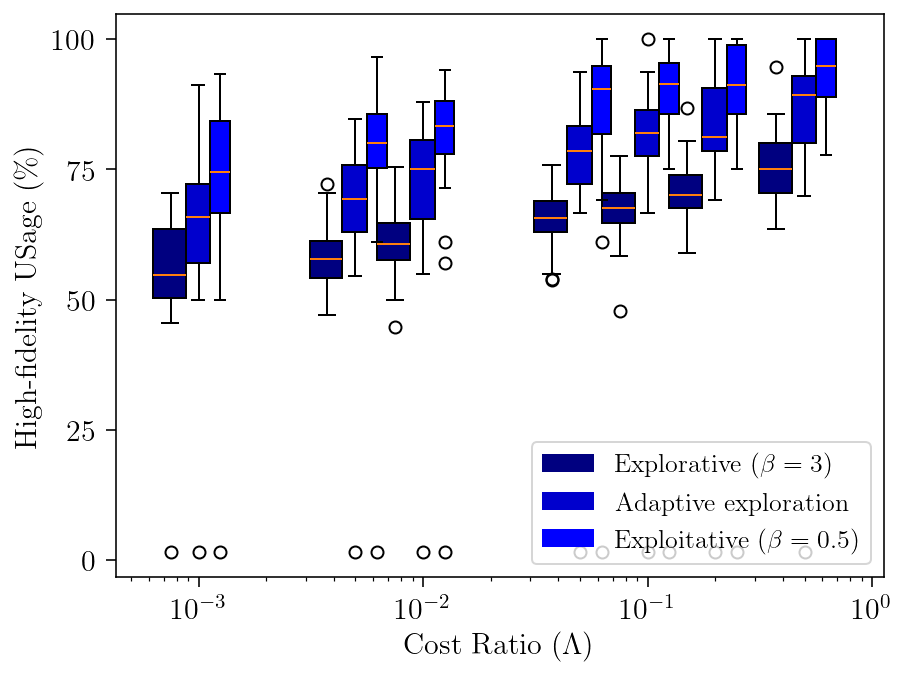}
     \end{subfigure}
     \hfill
     \begin{subfigure}[b]{0.32\textwidth}
         \centering
         \includegraphics[width=\textwidth]{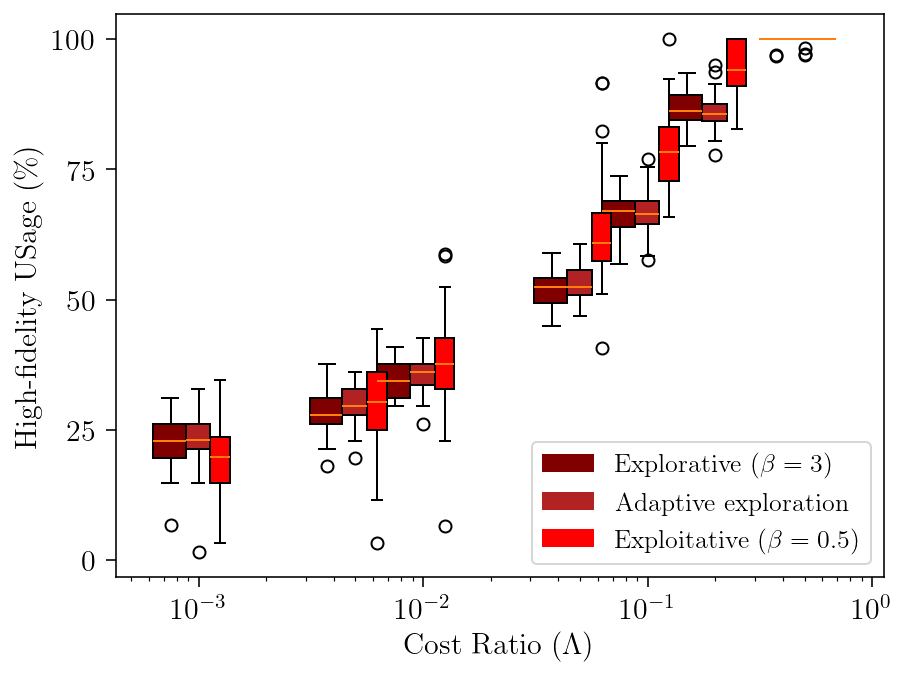}
     \end{subfigure}
     
     \caption{\textbf{Himmelblau 2D}. (a) Fidelity-weighted (b) MF-GPR-UCB (c) Proximity-based.}
     \label{fig:cost_himm}
     \end{figure}

\begin{figure}[H]
     \centering
     \begin{subfigure}[b]{0.32\textwidth}
         \centering
         \includegraphics[width=\textwidth]{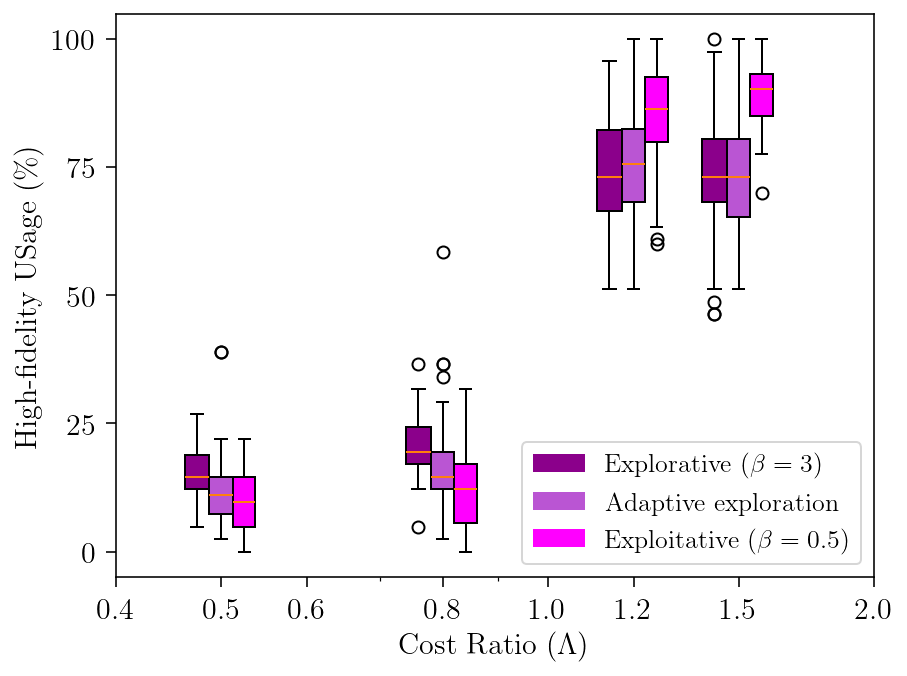}
     \end{subfigure}
     \hfill
     \begin{subfigure}[b]{0.32\textwidth}
         \centering
         \includegraphics[width=\textwidth]{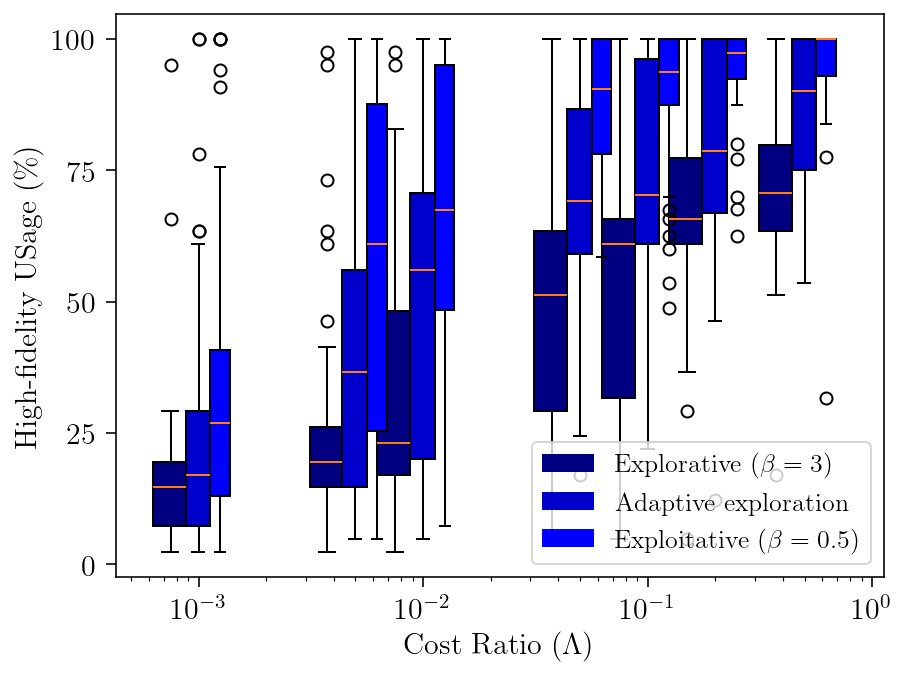}
     \end{subfigure}
     \hfill
     \begin{subfigure}[b]{0.32\textwidth}
         \centering
         \includegraphics[width=\textwidth]{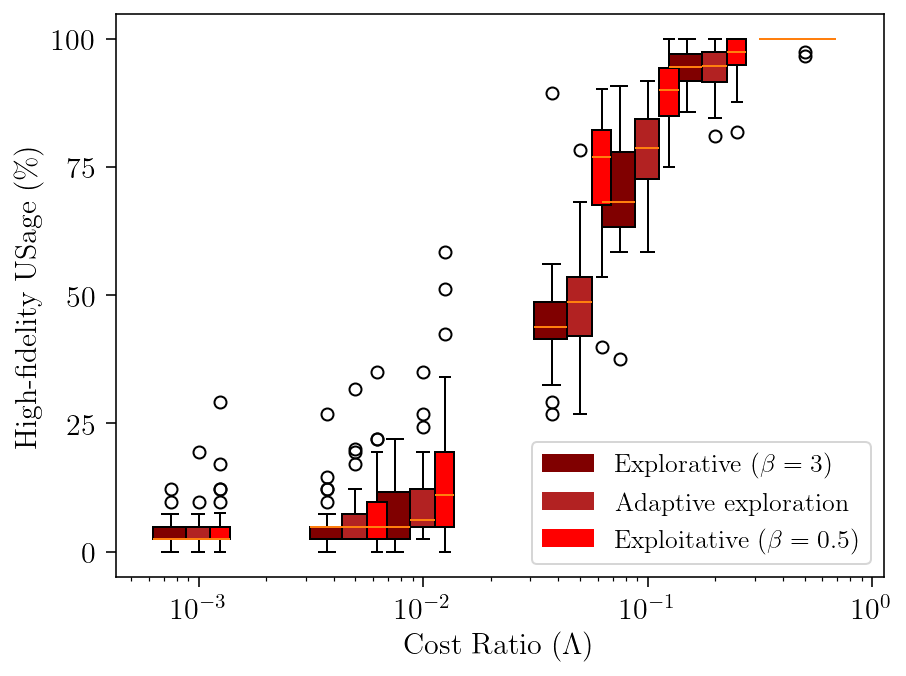}
     \end{subfigure}
     
     \caption{\textbf{Bohachevsky 2D}. (a) Fidelity-weighted (b) MF-GPR-UCB (c) Proximity-based. }
     \label{fig:cost_boh}
     \end{figure}

\section*{Acknowledgements}
\noindent This work was carried out at the Advanced Research Computing at Hopkins (ARCH) core facility (rockfish.jhu.edu), which is supported by the National Science Foundation (NSF) grant number OAC1920103. We also acknowledge Professors Dauenhauer and Vlachos for dynamic catalysis discussions, and especially Dr. G. Wittreich for providing us with his MATLAB ammonia catalysis model. 

\section*{Funding}
\noindent
AM, DG \& IGK: U.S. Department of Energy, Office of Science, Office of Advanced Scientific Computing Research grant under Award Number DE-SC0024162 and and by the National Science Foundation (NSF) under Grant 2436738\\
ASG \& IGK: US Air Force Office of Scientific Research (AFOSR MURI)

\bibliographystyle{elsarticle-num} 
\bibliography{biblio}

\end{document}